\pdfoutput=1
\documentclass[11pt]{article}

\usepackage{acl}
\usepackage[font=scriptsize]{subcaption}
\usepackage{times}
\usepackage{latexsym}
\usepackage{pifont}

\usepackage{subcaption}
\usepackage{arydshln}

\usepackage[utf8]{inputenc}
\usepackage{multirow}
\usepackage{microtype}
\usepackage{graphicx}
\usepackage{comment}
\usepackage{multicol}
\usepackage{booktabs}
\usepackage{xcolor}
\usepackage[font=small,labelfont=bf]{caption}
\definecolor{1}{RGB}{173, 24, 58}
\definecolor{2}{RGB}{178, 31, 60}
\definecolor{3}{RGB}{183, 38, 61}
\definecolor{4}{RGB}{188, 45, 63}
\definecolor{5}{RGB}{221, 101, 101}
\definecolor{6}{RGB}{225, 111, 111}
\definecolor{7}{RGB}{228, 121, 121}
\definecolor{8}{RGB}{232, 131, 131}
\definecolor{9}{RGB}{235, 141, 141}
\definecolor{10}{RGB}{238, 152, 152}
\definecolor{11}{RGB}{241, 162, 162}
\definecolor{12}{RGB}{243,173,172}
\definecolor{13}{RGB}{246,184,182}
\definecolor{14}{RGB}{248,195,192}
\definecolor{15}{RGB}{249,206,202}
\definecolor{16}{RGB}{251,216,212}
\definecolor{17}{RGB}{252,226,222}
\definecolor{18}{RGB}{253, 235, 232}
\definecolor{19}{RGB}{254,244, 242}
\definecolor{20}{RGB}{255, 253, 252}
\definecolor{d19}{RGB}{250,233, 176}

\definecolor{r7}{RGB}{132,159,219}
\definecolor{r8}{RGB}{140,164,223}
\definecolor{r9}{RGB}{148,169,226}
\definecolor{r10}{RGB}{157,174,230}
\definecolor{r11}{RGB}{166,180,233}
\definecolor{r12}{RGB}{175,187,236}
\definecolor{r13}{RGB}{184,193,239}
\definecolor{r14}{RGB}{193,200,242}
\definecolor{r15}{RGB}{202,208,244}
\definecolor{r16}{RGB}{212,216,247}
\definecolor{r17}{RGB}{222,224, 249}
\definecolor{r18}{RGB}{232,233,251}
\definecolor{r19}{RGB}{242,242, 253}
\definecolor{r20}{RGB}{252, 252, 255}

\definecolor{e7}{RGB}{104, 193, 109}
\definecolor{e8}{RGB}{117, 198, 115}
\definecolor{e9}{RGB}{134, 203, 126}
\definecolor{e10}{RGB}{151, 208, 137}
\definecolor{e11}{RGB}{166, 213, 148}
\definecolor{e12}{RGB}{180, 218, 159}
\definecolor{e13}{RGB}{194, 223, 171}
\definecolor{e14}{RGB}{206, 227, 182}
\definecolor{e15}{RGB}{217, 232, 194}
\definecolor{e16}{RGB}{227, 237, 205}
\definecolor{e17}{RGB}{236, 241, 217}
\definecolor{e18}{RGB}{243, 246, 229}
\definecolor{e19}{RGB}{249, 250, 241}
\definecolor{e20}{RGB}{254, 254, 253}

\definecolor{g1}{RGB}{192,199,197}
\definecolor{g2}{RGB}{205,210,209}
\definecolor{g3}{RGB}{217,221,220}
\definecolor{g4}{RGB}{230,233,232}
\definecolor{g5}{RGB}{242,244,243}

\definecolor{b1}{RGB}{244,250,255}
\definecolor{b2}{RGB}{232,248,255}
\definecolor{b3}{RGB}{200,234,255}
\definecolor{b4}{RGB}{178,230,255}
\definecolor{b5}{RGB}{156,203,255}

\definecolor{y0}{RGB}{255,251,220}
\definecolor{y1}{RGB}{255,251,200}
\definecolor{y2}{RGB}{255,250,160}
\definecolor{y3}{RGB}{255,241,134}
\definecolor{y4}{RGB}{255,229,118}
\definecolor{y5}{RGB}{255,217,96}

\title{Disentangling the Roles of Target-Side Transfer and Regularization \\ in Multilingual Machine Translation}

\author{
    Yan Meng \quad Christof Monz \\
    Language Technology Lab\\
    University of Amsterdam\\
    \texttt{\{y.meng, c.monz\}@uva.nl}
}
    
\begin{document}
\maketitle
\begin{abstract}

Multilingual Machine Translation (MMT) benefits from knowledge transfer across different language pairs. 
However, improvements in one-to-many translation compared to many-to-one translation are only marginal and sometimes even negligible.
This performance discrepancy raises the question of to what extent positive transfer plays a role on the target-side for one-to-many MT.
In this paper, we conduct a large-scale study that varies the auxiliary target-side languages along two dimensions, i.e., linguistic similarity and corpus size, to show the dynamic impact of knowledge transfer on the main language pairs. 
We show that linguistically similar auxiliary target languages exhibit strong ability to transfer positive knowledge. 
With an increasing size of similar target languages, the positive transfer is further enhanced to benefit the main language pairs. 
Meanwhile, we find distant auxiliary target languages can also unexpectedly benefit main language pairs, even with minimal positive transfer ability. 
Apart from transfer, we show distant auxiliary target languages can act as a regularizer to benefit translation performance by enhancing the generalization and model inference calibration.

\end{abstract}

\section{Introduction}

Multilingual Machine Translation (MMT) enables a single model to translate among multiple language pairs by joint training \citep{dong-etal-2015-multi, johnson-etal-2017-googles}. 
The improvements in translation quality, especially for low-resource languages, are generally attributed to transfer learning \citep{zoph-etal-2016-transfer, Lakew2018TransferLI, kocmi-bojar-2018-trivial, Stap2023ViewingKT}. 
However, MMT suffers from a performance gap where the gains for one-to-many translation are not as substantial as for many-to-one translation \citep{Dabre2020ACS,Tang2020MultilingualTW,Yang2021ImprovingMT,Chiang2021BreakingDM,Chowdhery2022PaLMSL}. 
Empirical studies \citep{johnson-etal-2017-googles, aharoni-etal-2019-massively} also show little to no benefit for one-to-many translation compared to their bilingual baselines, leading to the hypothesis that positive transfer does not occur on the target-side \citep{DBLP:journals/corr/massively}. 

The challenge of knowledge transfer for one-to-many translation is attributed to the inherent characteristics of translating into \emph{distinct} target languages. 
The necessity for target language-specific representations in the translation process hinders knowledge transfer as transfer learning prefers language-invariant representations \citep{kudugunta-etal-2019-investigating}.
On the other hand, \citet{DBLP:journals/corr/massively} and \citet{aharoni-etal-2019-massively} list the increasing amounts of source language data and regularization induced by multiple target languages as possible reasons for the observed benefits in massively MMT scenarios. 

Nevertheless, the extent to which positive knowledge transfer occurs on the target-side still remains unclear. 
Furthermore, a comprehensive analysis of the interplay between different factors, i.e., knowledge transfer, source data size, and regularization, for one-to-many translation is lacking. This hinders the optimization of MMT performance.   

To understand the impact of knowledge transfer, we conduct comprehensive controlled experiments with varying target languages along two dimensions, i.e., linguistic similarity and corpus size. 
We select a set of bilingual out-of-English translation tasks, e.g., English into German, as main language pairs. 
Subsequently, we add different auxiliary target language pairs to each main language pair, considering variations in auxiliary language families, written scripts, data sizes, and the number of target languages. 
Our experimental results show a consistent positive correlation between the improvements and their translation task relatedness, i.e., increasing the amounts of similar target languages encourages more positive knowledge transfer for the main language pair than distant ones. 
These findings confirm the existence of knowledge transfer on the target-side and also clearly show factors that influence target-side transfer, i.e., target data size, number of translation tasks, and linguistic similarity. 
The performance differences induced by various target languages also indicate their varying transfer ability.

Apart from knowledge transfer, we find distant auxiliary target languages can also yield substantial improvements, even with minimal transfer ability. 
Instead of transferring similar linguistic knowledge, we show that distant auxiliary target languages exhibit strong regularization abilities improving translation performance.% by reducing generalization errors and improving inference calibration. 
To understand why language regularization plays a role, we show it benefits translation performance by reducing generalization errors and improving inference calibration. 
With introducing distant auxiliary target languages, the translation model is implicitly calibrated so that the confidences of its predictions are more aligned with the accuracies of its predictions.

\begin{figure}[t!b]
    \centering
    \includegraphics[width=0.5\textwidth]{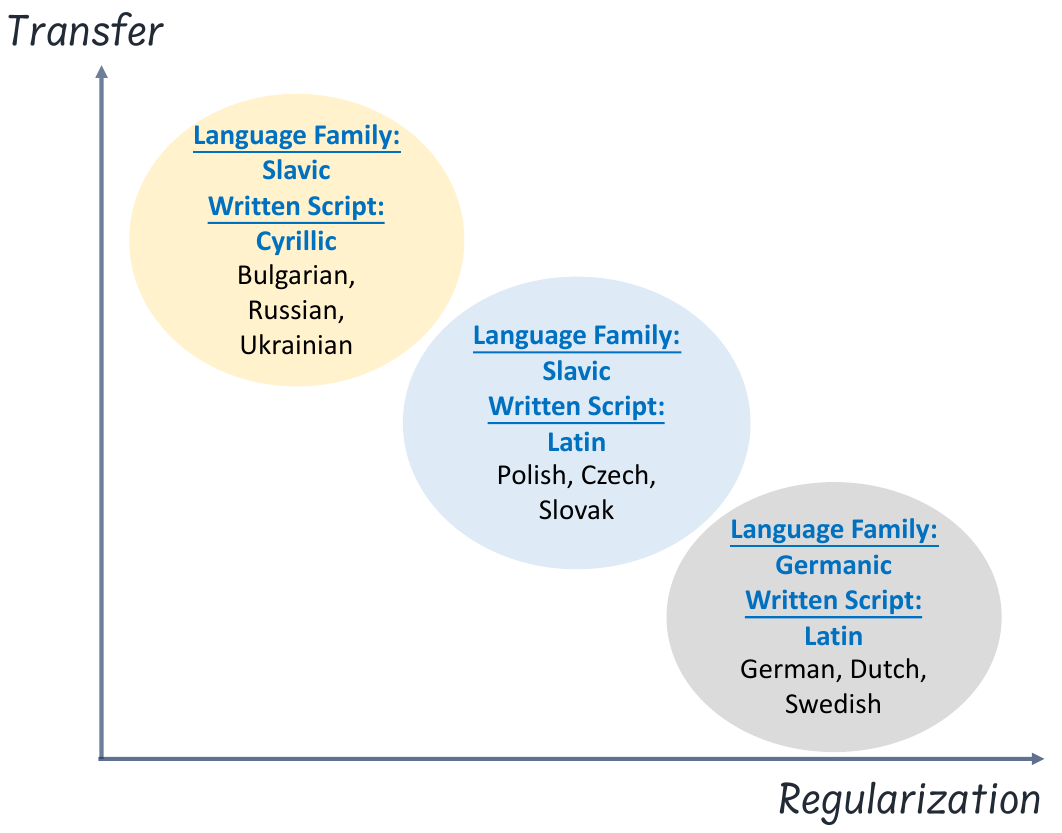}
    \caption{The interplay between knowledge transfer and regularization. For one example of main target language \textbf{Belarusian} (language family: Slavic, written script: Cyrillic) the level of knowledge transfer and regularization induced by different auxiliary target languages in MMT.}
    \label{fig:1}
   
\end{figure}

In this paper, we show the interplay between knowledge transfer and regularization which is visualized in Figure~\ref{fig:1}. We observe that languages that are similar to the target language, in this example \textit{Belarusian}, tend to benefit the target language by mostly transferring knowledge and only act as a regularizer to a very limited extent. The inverse holds for distant auxiliary languages. 
Overall, our paper provides a more comprehensive understanding of one-to-many translation, from the perspectives of target-side transfer and regularization.
First, we show how positive knowledge transfer occurs on the target-side, by varying the linguistic similarity and data size of the auxiliary target language. 
Second, we point out the importance of regularization in one-to-many translation, by showing its effectiveness on generalization ability and inference confidence calibration. 
% Furthermore, our findings provide new insights into different functions of auxiliary languages (as shown in Figure~\ref{fig:1}), i.e., transfer and regularization, which can be leveraged as simple and alternative methods to enhance real-world low-resource language pairs. 

\section{Background}
In this section, we discuss some background on transfer learning and regularization in MMT. 

\subsection{Transfer Learning}

Transfer learning is defined as improving a learner for a given task by leveraging information from a related task~\citep{RN4}.
An example is seen in MMT, where training models on multiple language pairs benefits resource-poor languages by leveraging shared linguistic information and parameters from other languages~\citep{zoph-etal-2016-transfer, murthy-etal-2019-addressing}.

However, for one-to-many machine translation, gains are much more pronounced for many-to-one  than for one-to-many translation.
This performance discrepancy is caused by the complexities of target-side transfer. 
\citet{aharoni-etal-2019-massively} empirically measure the difficulty of target-side transfer by showing the marginal benefits, even for low-resource language pairs, for large-scale one-to-many translation. 
\citet{Dabre2020ACS} suggest that the reason behind this challenge is mainly due to its characteristics of representations on the decoder side, where each target language has an independent output distribution and the decoder representations are more sensitive to the target languages \citep{kudugunta-etal-2019-investigating}.
\citet{Wang2018ThreeST} further supports this claim by keeping target language-specific parameters to improve one-to-many translation. 
This increases uncertainties on the effectiveness of transfer learning on the target-side, which in turn prefers language-invariant representations. 

Despite prior work \citep{gao-etal-2020-improving, mitigating} indicating that linguistic similarity matters for encouraging positive target-side transfer, their findings are limited to scenarios where knowledge is transferred from high-resource to low-resource. \citet{Fernandes2023ScalingLF} conversely shows that no impact of linguistic similarity on the translation performance for translating into two high-resource target languages, with an example of translating English into \{French, Chinese\} and English into \{French, German\}.

In summary, these studies show an inconsistent view towards target-side transfer, particularly concerning the issue whether target-side transfer exists and what factors influence it. This disagreement indicates the importance of exploring target-side transfer in one-to-many MT and the impact of different factors, e.g., linguistic similarity and target data size.

%Moreover, understanding the impact of various factors (e.g., linguistic similarity, target data size) on the transfer is also crucial. 

\subsection{Regularization}

The multilingual training regime is known as a source of regularization, improving the generalization abilities of the models \citep{neubig-hu-2018-rapid, aharoni-etal-2019-massively, Dabre2020ACS}. \citet{aharoni-etal-2019-massively} support this claim by showing that adding out-of-English translation tasks can lead to better results, as it prevents the model to overfit on the target side. 

However, the effects of language regularization induced by multiple target tasks are under-explored, compared to other regularization techniques, such as dropout~\citep{dropout} and label smoothing~\citep{Szegedy2015RethinkingTI}. 
Dropout randomly selects activations to be set to zero during training. This randomness introduced by dropout encourages the network to learn robust and generalized representations~\citep{Liang2021RDropRD}. 
Another common regularization technique, label smoothing, regularizes the model by penalizing the output confidence. 
It has also been shown that these changes in output confidence introduced by label smoothing could implicitly enhance machine translation model calibration~\citep{Mller2019WhenDL}, thereby improving translation performance. 
In line with this, we aim to investigate language regularization in one-to-many translation to understand when and why it is effective.

\section{Experimental Setting}

\paragraph{Model.}
We follow the setup of the Transformer base model~\cite{NIPS2017_3f5ee243}.
More details on model hyperparameters can be found in Appendix~\ref{sec:appendix-model}.

\paragraph{Data.} 

We choose three main language pairs (LPs) in different language families and written scripts: English-into-German (En$\rightarrow$De), English-into-Russian (En$\rightarrow$Ru), and English-into-Spanish (En$\rightarrow$Es).  
The training data for the main language pairs En$\rightarrow$De, En$\rightarrow$Ru, and En$\rightarrow$Es are from WMT13, WMT14, and WMT22, respectively. 
We randomly sample 100K and 1M sentence pairs from each language pair respectively to mimic low- and medium-resource settings\footnote{Using high-resource LPs to mimic low/med-resource LPs helps compare the transfer and regularization levels induced from the same and other target languages.}.
We also choose two real world low- and medium-resource language pairs: English-into-Belarusian (En$\rightarrow$Be) and English-into-Sinhala (En$\rightarrow$Si) from the OPUS repository.\footnote{https://opus.nlpl.eu}
%To observe the impact on high-resource settings, we use the full training corpus for En$\rightarrow$De (4.5M examples). 
For different controlled experiments, we cover 20 auxiliary target language pairs to train together with the main translation tasks. 
We randomly sample the auxiliary covered language pairs from CCMatrix. 
The detailed statistics of the main and auxiliary language pairs can be found in Appendix~\ref{sec:appendix-data}.

\begin{table}[!t]
\centering
    
   \resizebox{0.75\linewidth}{!}{
        \begin{tabular}{cc|cc}
         
        \toprule
           ISO & Lang. & Family & Script \\
              \midrule
            \textbf{De} & \textbf{German} &  \textbf{Germanic} & \textbf{Latin} \\
            \hdashline
            Nl & Dutch & Germanic & Latin \\
            Et & Estonia & Uralic & Latin \\
            Ru & Russian & Slavic & Cyrillic \\ 
            Zh & Mandarin &Chinese  & Chinese \\ 
            \bottomrule
            \textbf{ Es} & \textbf{Spanish} &\textbf{ Romance }&\textbf{ Latin} \\
            \hdashline
            Pt & Portuguese & Romance & Latin \\
            Nl &  Dutch & Germanic & Latin \\
            Ru &  Russian & Slavic & Cyrillic \\
            Zh & Mandarin &  Chinese  & Chinese \\ 
            \bottomrule
    
            \textbf{Ru} &\textbf{Russian} & \textbf{Slavic} &\textbf{ Cyrillic} \\
            \hdashline
            Uk &  Ukrainian &Slavic & Cyrillic \\
            Cs & Czech &  Slavic & Latin \\
            De & German &  Germanic & Latin \\
            Zh & Mandarin & Chinese  & Chinese \\ 
            
            \bottomrule
            \textbf{Si} &\textbf{Sinhala} & \textbf{Indo-Aryan} &\textbf{Sinhala} \\
            \hdashline
            Hi &  Hindi & Indo-Aryan  & Devanagari  \\
            Ur & Urdu &  Indo-Aryan & Arabic \\
            De & German &  Germanic & Latin \\
            Zh & Mandarin & Chinese  & Chinese \\ 
            \bottomrule
            \textbf{Be} &\textbf{Belarusian} & \textbf{Slavic} &\textbf{Cyrillic} \\
            \hdashline
            Ru &  Russian & Slavic &  Cyrillic \\
            Pl &  Polish &  Slavic & Latin \\
            De & German &  Germanic & Latin \\
            Zh & Mandarin & Chinese  & Chinese \\ 
            \bottomrule
            \end{tabular}}
        
    \caption{Linguistic information for the main and auxiliary target languages.  \textbf{Bold} designates the main target languages: De, Es, Ru, Si, and Be.  }
    \label{tab:liguistic}
    
\end{table}

\begin{table*}[!t]
    \centering
    \begin{subtable}{\linewidth}
   \resizebox{\linewidth}{!}{
    \begin{tabular}{r|r:rrrr}
    \toprule
        \multicolumn{6}{c}{\textbf{En$\rightarrow$De (Baseline: 7.4)} }\\
    
    $\alpha\%$ & en$\rightarrow$de   & en$\rightarrow$nl & en$\rightarrow$et & en$\rightarrow$ru & en$\rightarrow$zh \\
    \midrule
       10\% & 8.5\textsubscript{0.4} & \colorbox{20}{7.9\textsubscript{0.7}} &  \colorbox{19}{8.2\textsubscript{0.6}} & \colorbox{18}{8.6\textsubscript{0.5}} & \colorbox{18}{8.9\textsubscript{0.8}} \\
       50\% & {10.2\textsubscript{0.3}} & \colorbox{18}{10.3\textsubscript{0.6}}  & \colorbox{18}{10.5\textsubscript{0.6}}  & \colorbox{17}{10.9\textsubscript{0.3}} & \colorbox{16}{{11.5}\textsubscript{0.4}} \\
       100\% & {11.6\textsubscript{0.4}} & \colorbox{16}{11.3\textsubscript{0.4}}  & \colorbox{17}{10.9\textsubscript{0.2}} & \colorbox{16}{11.0\textsubscript{0.4}} & \colorbox{14}{{12.1}\textsubscript{0.2}} \\
       500\% & {15.9\textsubscript{0.3}}  &\colorbox{12}{14.0\textsubscript{0.2}}& \colorbox{12}{{13.7}\textsubscript{0.3}}   &  \colorbox{13}{13.4\textsubscript{0.2}} & \colorbox{13}{13.5\textsubscript{0.3}}\\
       1000\%  & {19.9\textsubscript{0.1}} & \colorbox{7}{16.2\textsubscript{0.2}} & \colorbox{9}{15.3\textsubscript{0.1}} & \colorbox{12}{14.1\textsubscript{0.2}} & \colorbox{12}{14.2\textsubscript{0.1}} \\
    \bottomrule

    \end{tabular}

    \begin{tabular}{r|r:rrrr}
    \toprule
        \multicolumn{6}{c}{\textbf{En$\rightarrow$De (Baseline: 20.0)} }\\
    
    $\alpha\%$ & en$\rightarrow$de & en$\rightarrow$nl & en$\rightarrow$et & en$\rightarrow$ru & en$\rightarrow$zh \\
    \midrule
       1\% & 20.0\textsubscript{0.4} & \colorbox{20}{20.2\textsubscript{0.4}} &  \colorbox{19}{20.5\textsubscript{0.2}} & \colorbox{17}{20.7\textsubscript{0.3}} & \colorbox{17}{20.8\textsubscript{0.5}}   \\
       10\% & 20.3\textsubscript{0.2} & \colorbox{16}{21.0\textsubscript{0.3}}  & \colorbox{17}{20.7\textsubscript{0.4}}  & \colorbox{12}{21.2\textsubscript{0.6}} & \colorbox{10}{21.8\textsubscript{0.6}} \\
       50\% &  22.1\textsubscript{0.4} &\colorbox{15}{21.6\textsubscript{0.5}}  & \colorbox{14}{21.3\textsubscript{0.1}}  & \colorbox{14}{21.2\textsubscript{0.2}} & \colorbox{14}{21.6\textsubscript{0.2}} \\
       100\% &  23.4\textsubscript{0.2} &\colorbox{10}{22.2\textsubscript{0.2}}& \colorbox{15}{21.2\textsubscript{0.2}}   &  \colorbox{15}{21.0\textsubscript{0.2}} & \colorbox{15}{21.2\textsubscript{0.2}}\\
       200\% & 24.5\textsubscript{0.1} & \colorbox{12}{22.2\textsubscript{0.0}} & \colorbox{19}{20.2\textsubscript{0.0}} & \colorbox{20}{20.0\textsubscript{0.0}} & \colorbox{19}{20.7\textsubscript{0.0}} \\
    \bottomrule

    \end{tabular}}
    
       \end{subtable}

\begin{subtable}{\linewidth}
   \resizebox{\linewidth}{!}{
   \begin{tabular}{r|c:cccc}
    \toprule
        \multicolumn{6}{c}{\textbf{En$\rightarrow$Ru (Baseline: 11.9) } }\\
    
    $\alpha\%$& en$\rightarrow$ru   & en$\rightarrow$uk & en$\rightarrow$cs & en$\rightarrow$de & en$\rightarrow$zh \\
    \midrule
       10\%  & 12.0\textsubscript{0.4} & \colorbox{r20}{11.8\textsubscript{0.6}} &  \colorbox{r20}{11.6\textsubscript{0.6}} & \colorbox{r20}{11.7\textsubscript{0.2}} & \colorbox{r19}{{12.0}\textsubscript{0.4}}   \\
       50\%  & 12.8\textsubscript{0.3} & \colorbox{r18}{13.0\textsubscript{0.5}}  & \colorbox{r19}{{12.2}\textsubscript{0.2}}  & \colorbox{r18}{12.4\textsubscript{0.3}} & \colorbox{r18}{12.6\textsubscript{0.1}} \\
       100\% & 14.0\textsubscript{0.2} & \colorbox{r17}{13.3\textsubscript{0.3}}  & \colorbox{r18}{12.6\textsubscript{0.1}}  & \colorbox{r18}{12.7\textsubscript{0.2}} &\colorbox{r18}{12.8\textsubscript{0.4}} \\
       500\% & 15.7\textsubscript{0.2} & \colorbox{r14}{14.7\textsubscript{0.2}}& \colorbox{r16}{{14.2}\textsubscript{0.1}}   &  \colorbox{r15}{14.4\textsubscript{0.2}} & \colorbox{r14}{14.6\textsubscript{0.1}}\\
       1000\%  & 18.6\textsubscript{0.3} & \colorbox{r11}{15.4\textsubscript{0.1}} & \colorbox{r13}{{14.7}\textsubscript{0.2}} & \colorbox{r15}{14.6\textsubscript{0.2}} & \colorbox{r16}{14.3\textsubscript{0.2}} \\
    \bottomrule

    \end{tabular}
    
 \begin{tabular}{r|r:rrrr}
    \toprule
        \multicolumn{6}{c}{\textbf{En$\rightarrow$Ru (Baseline: 18.4)} }\\
    
    $\alpha\%$  & en$\rightarrow$ru & en$\rightarrow$uk & en$\rightarrow$cs & en$\rightarrow$de & en$\rightarrow$zh \\
    \midrule
       1\%  & 18.1\textsubscript{0.3} & \colorbox{r20}{18.6\textsubscript{0.5}} &  \colorbox{r20}{18.7\textsubscript{0.8}} & \colorbox{r20}{18.7\textsubscript{0.5}} & \colorbox{r18}{18.9\textsubscript{0.2}}   \\
       10\%  & 18.6\textsubscript{0.5} & \colorbox{r18}{18.9\textsubscript{0.2}}  & \colorbox{r16}{19.1\textsubscript{0.1}}  & \colorbox{r16}{18.9\textsubscript{0.2}} & \colorbox{r16}{19.1\textsubscript{0.3}} \\
       50\% & 19.5\textsubscript{0.2} & \colorbox{r16}{19.3\textsubscript{0.3}}  & \colorbox{r17}{18.8\textsubscript{0.1}}  & \colorbox{r18}{18.4\textsubscript{0.2}} & \colorbox{r17}{18.7\textsubscript{0.1}} \\
       100\% & 20.1\textsubscript{0.1} & \colorbox{r16}{19.5\textsubscript{0.2}}& \colorbox{r17}{19.1\textsubscript{0.1}}  &  \colorbox{r18}{18.6\textsubscript{0.2}} & \colorbox{r19}{18.2\textsubscript{0.1}}\\
       200\%  &22.4\textsubscript{0.1} & \colorbox{r13}{20.5\textsubscript{0.0}} & \colorbox{r19}{18.5\textsubscript{0.0}} & \colorbox{r20}{17.2\textsubscript{0.0}} & \colorbox{r20}{17.1\textsubscript{0.0}} \\
    \bottomrule

    \end{tabular}}
    \end{subtable}

    \begin{subtable}{\linewidth}
\resizebox{\linewidth}{!}{
    \begin{tabular}{r|l:cccc}
    \toprule
        \multicolumn{6}{c}{\textbf{En$\rightarrow$Es (Baseline: 16.9)} }\\
  %  & \textbf{main} & \multicolumn{4}{c}{\textbf{auxilliary}}\\
    $\alpha\%$ & en$\rightarrow$es   & en$\rightarrow$pt & en$\rightarrow$nl & en$\rightarrow$ru & en$\rightarrow$zh \\
    \midrule
    %   0\% & 16.9\textsubscript{\ \ \ \ \ \ } \\
       10\% & 17.1\textsubscript{0.2} & \colorbox{e20}{17.0\textsubscript{0.4}} &  \colorbox{e19}{17.3\textsubscript{0.6}} & \colorbox{e19}{17.2\textsubscript{0.3}} & \colorbox{e18}{17.6\textsubscript{0.8}}   \\
       50\% & {19.0\textsubscript{0.2}} & \colorbox{e19}{18.1\textsubscript{0.3}}  & \colorbox{e17}{18.5\textsubscript{0.6}}  & \colorbox{e16}{19.0\textsubscript{0.2}} & \colorbox{e15}{{19.5}\textsubscript{0.3}} \\
       100\% & {20.9\textsubscript{0.4}} & \colorbox{e16}{19.1\textsubscript{0.3}}  & \colorbox{e15}{19.4\textsubscript{0.3}} & \colorbox{e16}{19.1\textsubscript{0.3}} & \colorbox{e14}{{21.0}\textsubscript{0.2}} \\
       500\% & {27.1\textsubscript{0.3}}  &\colorbox{e12}{23.2\textsubscript{0.2}}& \colorbox{e13}{{21.5}\textsubscript{0.3}}   &  \colorbox{e13}{22.8\textsubscript{0.3}} & \colorbox{e13}{23.0\textsubscript{0.2}}\\
       1000\%  & {29.4\textsubscript{0.2}} & \colorbox{e7}{25.2\textsubscript{0.4}} & \colorbox{e10}{23.2\textsubscript{0.1}} & \colorbox{e11}{22.4\textsubscript{0.3}} & \colorbox{e11}{22.2\textsubscript{0.1}} \\
    \bottomrule

    \end{tabular}
    \begin{tabular}{r|r:rrrr}
    \toprule
        \multicolumn{6}{c}{\textbf{En$\rightarrow$Es (Baseline: 28.6)} }\\
    
    $\alpha\%$ & en$\rightarrow$es & en$\rightarrow$pt & en$\rightarrow$nl & en$\rightarrow$ru & en$\rightarrow$zh \\
    \midrule
       1\% & 28.6\textsubscript{0.3} & \colorbox{e19}{28.6\textsubscript{0.1}} &  \colorbox{e19}{28.7\textsubscript{0.2}} & \colorbox{e19}{28.8\textsubscript{0.2}} & \colorbox{e19}{28.7\textsubscript{0.5}}   \\
       10\% & 29.4\textsubscript{0.2} & \colorbox{e17}{29.0\textsubscript{0.3}}  & \colorbox{e17}{29.1\textsubscript{0.2}}  & \colorbox{e13}{29.3\textsubscript{0.4}} & \colorbox{e13}{29.2\textsubscript{0.3}} \\
       50\% &  29.9\textsubscript{0.4} &\colorbox{e15}{29.2\textsubscript{0.5}}  & \colorbox{e12}{29.4\textsubscript{0.2}}  & \colorbox{e12}{29.4\textsubscript{0.2}} & \colorbox{e12}{29.4\textsubscript{0.1}} \\
       100\% &  30.5\textsubscript{0.3} &\colorbox{e12}{29.5\textsubscript{0.3}}& \colorbox{e13}{29.2\textsubscript{0.1}}   &  \colorbox{e17}{29.0\textsubscript{0.3}} & \colorbox{e13}{29.2\textsubscript{0.4}}\\
       200\% & 31.8\textsubscript{0.2} & \colorbox{e11}{29.6\textsubscript{0.0}} & \colorbox{e19}{28.9\textsubscript{0.0}} & \colorbox{e20}{28.3\textsubscript{0.0}} & \colorbox{e19}{28.0\textsubscript{0.0}} \\
    \bottomrule

    \end{tabular}}
    \end{subtable}

    \caption{BLEU scores (variance in subscript) for the three main tasks: En$\rightarrow$De, En$\rightarrow$Es, and En$\rightarrow$Ru in low-resource 100K (left) and medium-resource 1M (right) settings when training with different auxiliary language pairs. $\alpha\%$ represents the auxiliary training data size. For low-resource setting, $\alpha\%$ ranges from 10\% to 1000\% of the proportion of the low-resource setting size. For the medium-resource setting, $\alpha\%$ ranges from 1\% to 200\% of the proportion of the medium-resource setting size. The color block represents the extent of positive transfer, with darker shades indicating a stronger positive transfer effect. }
    \label{tab:low-medium-transfer}

\end{table*}

\paragraph{Training and Evaluation.}

We use the Fairseq~\citep{ott-etal-2019-fairseq} toolkit to train transformer models. 
All models are trained with the Adam optimizer~\citep{kingma2017adam} for up to 100K steps, with a learning rate of 5e-4 and an inverse square root scheduler. A dropout rate of 0.3 and label smoothing of 0.2 are used.  
Each model is trained on one NVIDIA A6000 GPU with a batch size of 25K tokens. 
We choose the best checkpoint according to the average validation loss of all language pairs. 
The data is tokenized with the SentencePiece tool~\citep{kudo2018sentencepiece} and we build a shared vocabulary of 32K tokens. 
We add language ID tokens to the vocabulary and prepend the language ID token to each source and target sequence to indicate the target language~\citep{johnson-etal-2017-googles}. 
For evaluation, we employ beam search decoding with a beam size of 5. 
BLEU scores are computed using detokenized case-sensitive SacreBLEU\footnote{nrefs:1$|$case:mixed$|$eff:no$|$tok:13a$|$smooth:exp$|$version:2.3.1}~\citep{post-2018-call}. 

\section{Target-Side Transfer}\label{sec:4}

In this section, we aim to estimate empirically whether target-side transfer occurs in MMT. 
To achieve this, we select three main language pairs, mimicking a low-resource direction: En$\rightarrow$De, En$\rightarrow$Es, En$\rightarrow$Ru and two main real-world low-resource pairs: En$\rightarrow$Be and En$\rightarrow$Si. 
We train each main language pair with different auxiliary target languages to investigate the target-side transfer in multilingual machine translation for influencing main language pairs.  
We include variations in the auxiliary target language pairs, with changes in linguistic similarity, data size, and the total number of target tasks. 

%We train each main language pair with different auxiliary target languages to investigate the target-side transfer in multilingual machine translation for influencing main language pairs.  
%We make variations in the auxiliary target language pairs, with changes in linguistic similarity, data size, and the total target task number. 

%The average results are reported along with the corresponding variance (Table~\ref{tab:low-resource} and Table~\ref{tab:mid-resource}). 

\subsection{Changes in Target Language}\label{sec:4.1} Here, we introduce different auxiliary target languages with variations in linguistic similarity and data size. 
The varying auxiliary target data size represents the true distribution of varied data in multilingual machine translation. 

\subsubsection{Setup} For each main language pair (En$\rightarrow$X), we train it with an auxiliary language pair (En$\rightarrow$Y) that differs in language family and written script. 
Table~\ref{tab:liguistic} presents the linguistic information about the main and auxiliary target languages. 
For the auxiliary target data, which is trained jointly with the main low-resource language pair, we vary its data size with a proportion from 10\% to 1000\% of the main low-resource language pair. 
For the auxiliary target data, trained jointly with the medium-resource setting, we vary its data size with a proportion from 1\% to 200\% of the main language pair.
To mitigate the variance in the quality of sampled auxiliary target language pairs, we run the experiment with three different randomly sampled sets.\footnote{We use one random sample set for high-resource (2M) auxiliary data due to computational constraints. }
Table~\ref{tab:low-medium-transfer} and Table~\ref{tab:real-case} show the averaged results of main mimic and real-world low- and medium-resource translation tasks when training with different target languages, along with the corresponding variance.

\begin{table}[!t]
    \centering
    \begin{subtable}{0.85\linewidth}
    \centering
   \resizebox{\linewidth}{!}{
    \begin{tabular}{r|rrrr}
    \toprule
        \multicolumn{5}{c}{\textbf{En$\rightarrow$Be (Baseline: 5.0)} }\\
    
    $\alpha\%$   & en$\rightarrow$ru & en$\rightarrow$pl & en$\rightarrow$de & en$\rightarrow$zh\\
    \midrule
       10\%  & \colorbox{y1}{    5.3\textsubscript{0.2}  } &  \colorbox{y1}{  
   5.1\textsubscript{0.1}  } & \colorbox{y0}{    4.2\textsubscript{0.3}  } & \colorbox{y0}{    4.3\textsubscript{0.3}  } \\
       50\% & \colorbox{y2}{    6.6\textsubscript{0.1}  }  & \colorbox{y0}{ 
   4.7\textsubscript{0.3}  }  & \colorbox{y1}{{    5.4\textsubscript{0.3}}  } & \colorbox{y1}{     5.8\textsubscript{0.2}  } \\
       100\% & \colorbox{y3}{    8.3\textsubscript{0.1}  }   & \colorbox{y1}{ 
   5.4\textsubscript{0.1}  } & \colorbox{y2}{{     6.2\textsubscript{0.2}  }} & \colorbox{y2}{{     7.0\textsubscript{0.2}  }} \\
       500\% &\colorbox{y5}{13.0\textsubscript{0.3}}&   \colorbox{y3}{  
   9.9\textsubscript{0.3}  } & \colorbox{y4}{10.0\textsubscript{0.4}} & \colorbox{y4}{11.0\textsubscript{0.2}}\\
       1000\%  & \colorbox{y5}{13.0\textsubscript{0.1}} & \colorbox{y3}{  
   9.4\textsubscript{0.3}  } & \colorbox{y3}{    9.3\textsubscript{0.2}  }   & \colorbox{y3}{10.0\textsubscript{0.3}}\\
    \bottomrule

    \end{tabular}}
 \end{subtable}
\begin{subtable}{0.85\linewidth}
\centering
\resizebox{\linewidth}{!}{
    \begin{tabular}{r|rrrr}
    \toprule
        \multicolumn{5}{c}{\textbf{En$\rightarrow$Si (Baseline: 22.6)} }\\
    
    $\alpha\%$ & en$\rightarrow$hi & en$\rightarrow$ur & en$\rightarrow$de  & en$\rightarrow$zh \\
    \midrule
       1\%  & \colorbox{y0}{22.8\textsubscript{0.2}} &  \colorbox{y2}{23.2\textsubscript{0.1}} & \colorbox{y0}{22.4\textsubscript{0.3}}& \colorbox{y1}{22.9\textsubscript{0.4}}   \\
       10\%  & \colorbox{y2}{23.2\textsubscript{0.2}}  & \colorbox{y1}{22.4\textsubscript{0.1}}  & \colorbox{y5}{23.9\textsubscript{0.3}} & \colorbox{y5}{24.0\textsubscript{0.3}} \\
       50\%  &\colorbox{y2}{23.3\textsubscript{0.2}}  & \colorbox{y1}{21.8\textsubscript{0.1}}  & \colorbox{y3}{23.5\textsubscript{0.4}} & \colorbox{y4}{23.7\textsubscript{0.3}} \\
       100\% & \colorbox{y5}{23.9\textsubscript{0.3}}& \colorbox{y1}{21.6\textsubscript{0.2}}   &  \colorbox{y3}{23.1\textsubscript{0.2}} & \colorbox{y3}{23.4\textsubscript{0.1}}\\
       200\% &  \colorbox{y4}{23.6\textsubscript{0.0}} & \colorbox{y1}{21.0\textsubscript{0.0}} & \colorbox{y1}{22.5\textsubscript{0.0}} & \colorbox{y1}{22.4\textsubscript{0.0}} \\
    \bottomrule

    \end{tabular}}
    
       \end{subtable}
       
    \caption{BLEU scores for the real-world low-resource English$\rightarrow$Belarusian (67K) and medium-resource English$\rightarrow$Sinhala (970K) from OPUS dataset. }
    \label{tab:real-case}

    \end{table}

\subsubsection{Discussion} \label{sec4.discus}

First, we show positive knowledge transfer occurs on the target-side, which benefits low- and medium-resource language pairs. 
This positive target-side transfer is highly correlated with translation task relatedness, i.e., linguistic similarity. 
Specifically, for low- and medium-resource settings, see Table~\ref{tab:low-medium-transfer}, increasing the amounts of similar target languages improves positive knowledge transfer for the main language pairs, i.e., 9 BLEU points boost for the low-resource En$\rightarrow$De task when training with 1000\% En$\rightarrow$Nl. 
However, training with the same amounts of a distant target task cannot achieve similar improvements, such as En$\rightarrow$Zh.
This also holds for the real-world low-resource En$\rightarrow$Be task, shown in Table~\ref{tab:real-case}. 
Increasing the size of a similar translation task, En$\rightarrow$Ru induces more positive knowledge transfer than other language pairs. 
Furthermore, the varying performance for the main tasks when training with different target-side languages shows that increasing the amount of English source data \citep{DBLP:journals/corr/massively} cannot be entirely confirmed as the sole reason for the improvements. 

%Although \citet{DBLP:journals/corr/massively} indicates another possible factor, i.e., increasing the source data, our varying translation performance for the main tasks when training with different target tasks cannot entirely confirm it as the sole reason. 

Second, we demonstrate that negative transfer also exists with increasing amounts of target data.
For medium-resource settings, increasing the size of distant auxiliary languages gradually shows the negative transfer for medium-resource main language pairs.
Training with 200\% of English to Chinese data leads to approximately 1.5 BLEU points drop for medium-resource English to Russian. 
%For the high-resource setting (Table \ref{tab:high-resource}), negative transfer almost occurs in training with every auxiliary language pair. 
This still correlates with linguistic similarity where distant data results in more performance drops than similar ones. 
In line with \citet{negative-transfer}, the divergence between joint distributions of tasks is the root of the negative transfer.

Third, we find that the gains for low- or medium-resource tasks in one-to-many translation cannot be fully attributed to transfer learning. 
Distant target languages which exhibit minimal positive transfer ability can also greatly improve the translation performance of the main language pairs. 
This becomes more evident when using small amounts of distant auxiliary languages. 
In Table~\ref{tab:low-medium-transfer} (right), joint training with 10\% distant language pairs can even lead to better translation performance for all main language tasks than using 10\% of similar data. 
10\% of En$\rightarrow$Zh data can even lead to an improvement of about 2 BLEU points for the En$\rightarrow$De task in a medium-resource setting. 
In the real-world medium-resource En$\rightarrow$Si task, training with 10\% of distant data En$\rightarrow$De or En$\rightarrow$Zh can outperform the maximum positive transfer induced by 200\% of similar language En$\rightarrow$Hi. 
The gains resulting from the small size of distant auxiliary data show the role of language regularization, discussed in Section~\ref{sec:5}.
By joint training with auxiliary low-resource target tasks, uncertainties are increased for the model to prevent over-fitting on the main tasks. 
Moreover, the unexpected benefits from distant auxiliary data on multilingual machine translation also encourages future work to exploit the role of distant data in other cross-lingual tasks. 
%Further discussion is shown in Section~\ref{sec:5}. 

\subsection{Changes in Task Number}\label{sec:4.2}

To further validate the previous findings, we expand the scenario from training with a single target task to incorporating multiple tasks. 
We control the total amount of auxiliary training data to ensure a fair comparison.

\subsubsection{Setup} 
\begin{figure}[!t]
    \centering
    \begin{subfigure}{0.23\textwidth}
        \includegraphics[width=\textwidth]{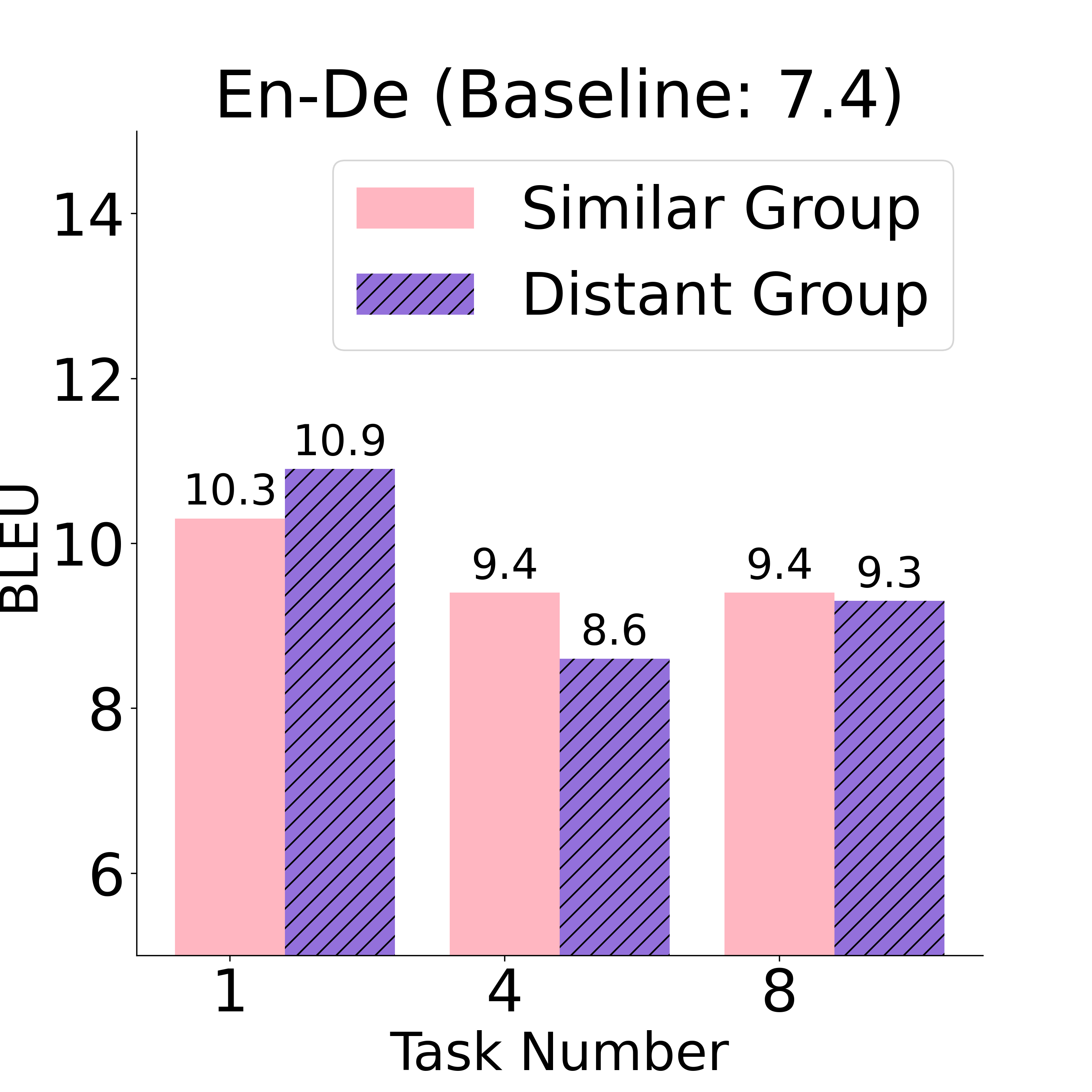}
        \subcaption{Data size: 50\%}
    \end{subfigure}
    \begin{subfigure}{0.23\textwidth}
        \includegraphics[width=\textwidth]{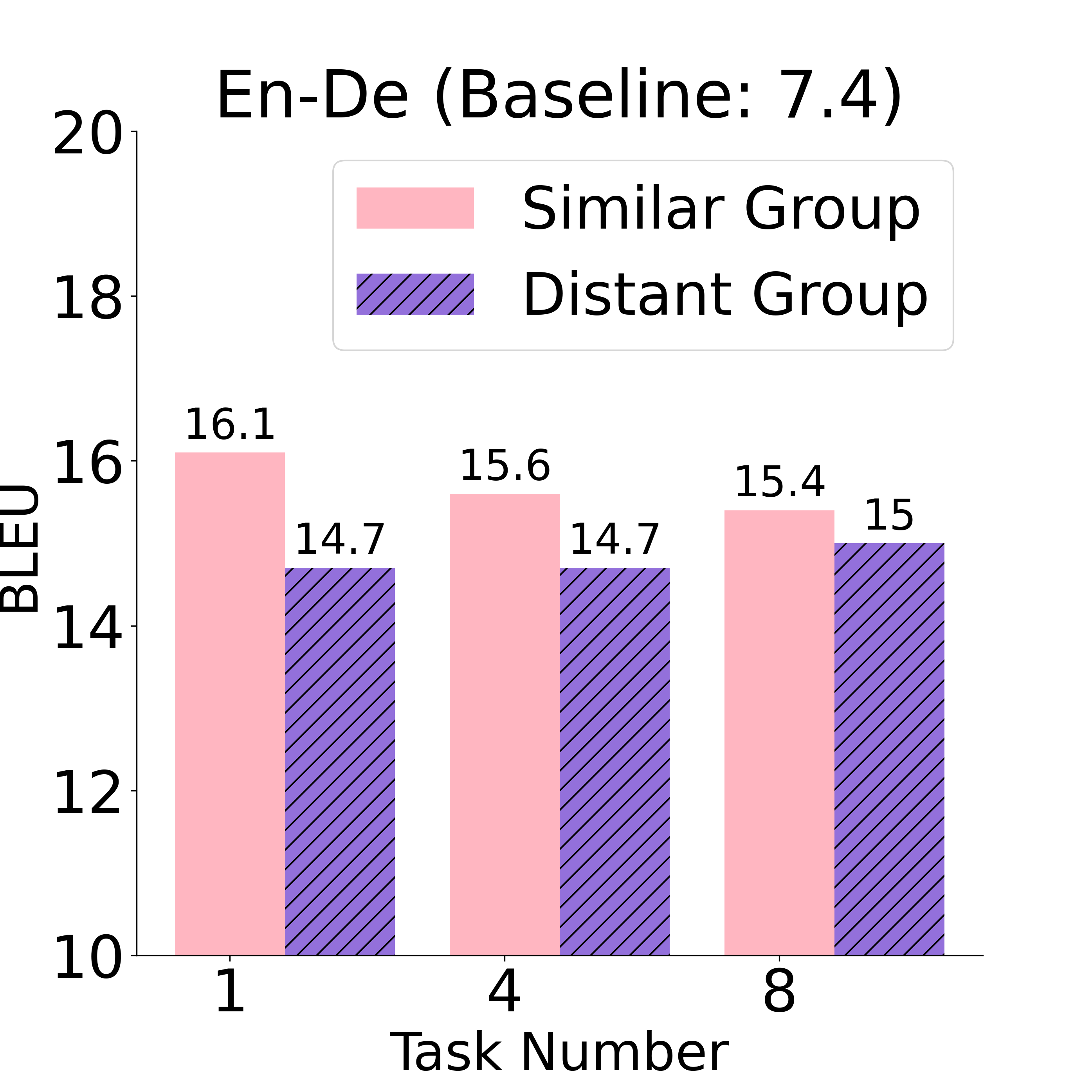}
        \subcaption{Data size: 1000\%}
    \end{subfigure}
    \begin{subfigure}{0.23\textwidth}
        \includegraphics[width=\textwidth]{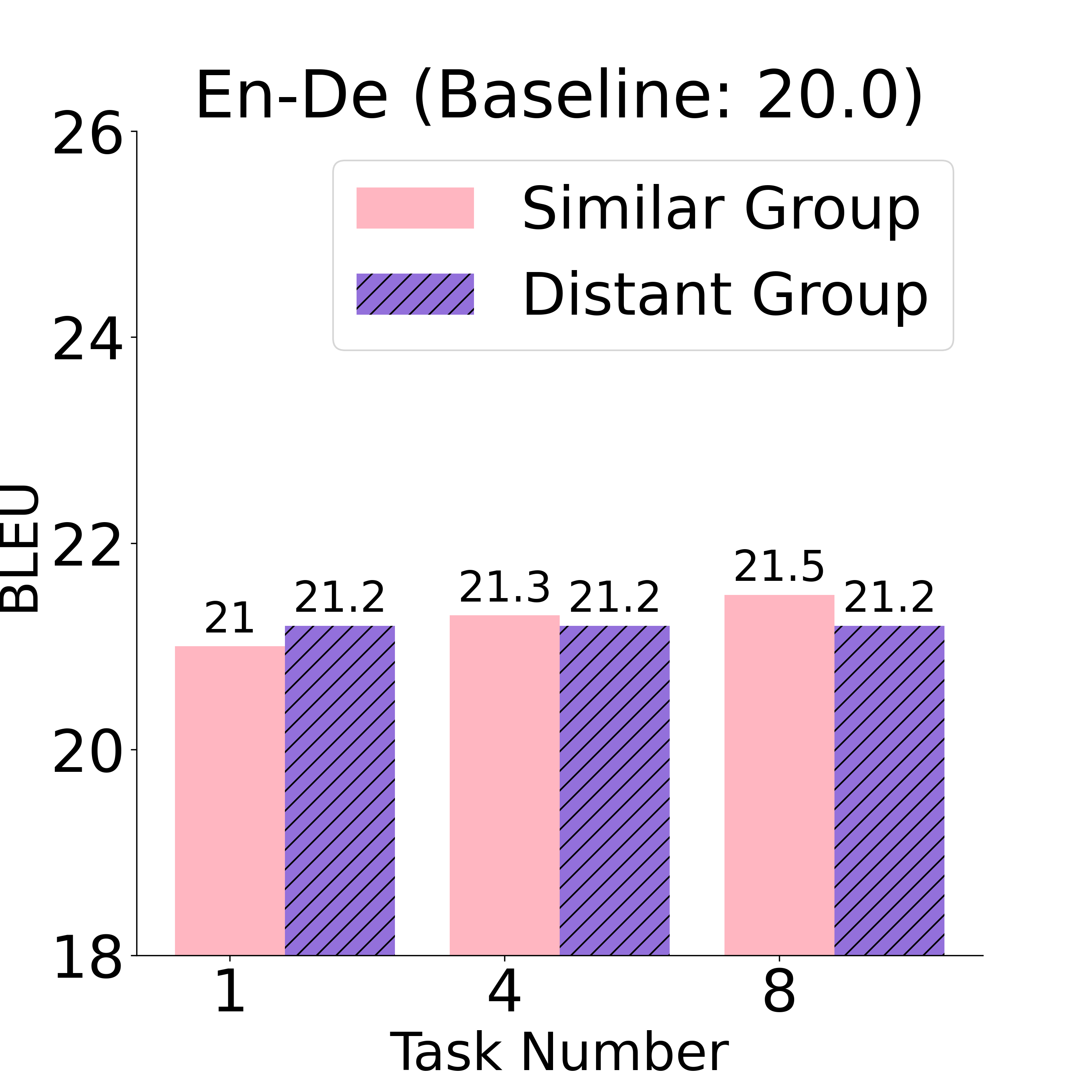}
        \subcaption{Data size: 10\%}
    \end{subfigure}
    \begin{subfigure}{0.23\textwidth}
        \includegraphics[width=\textwidth]{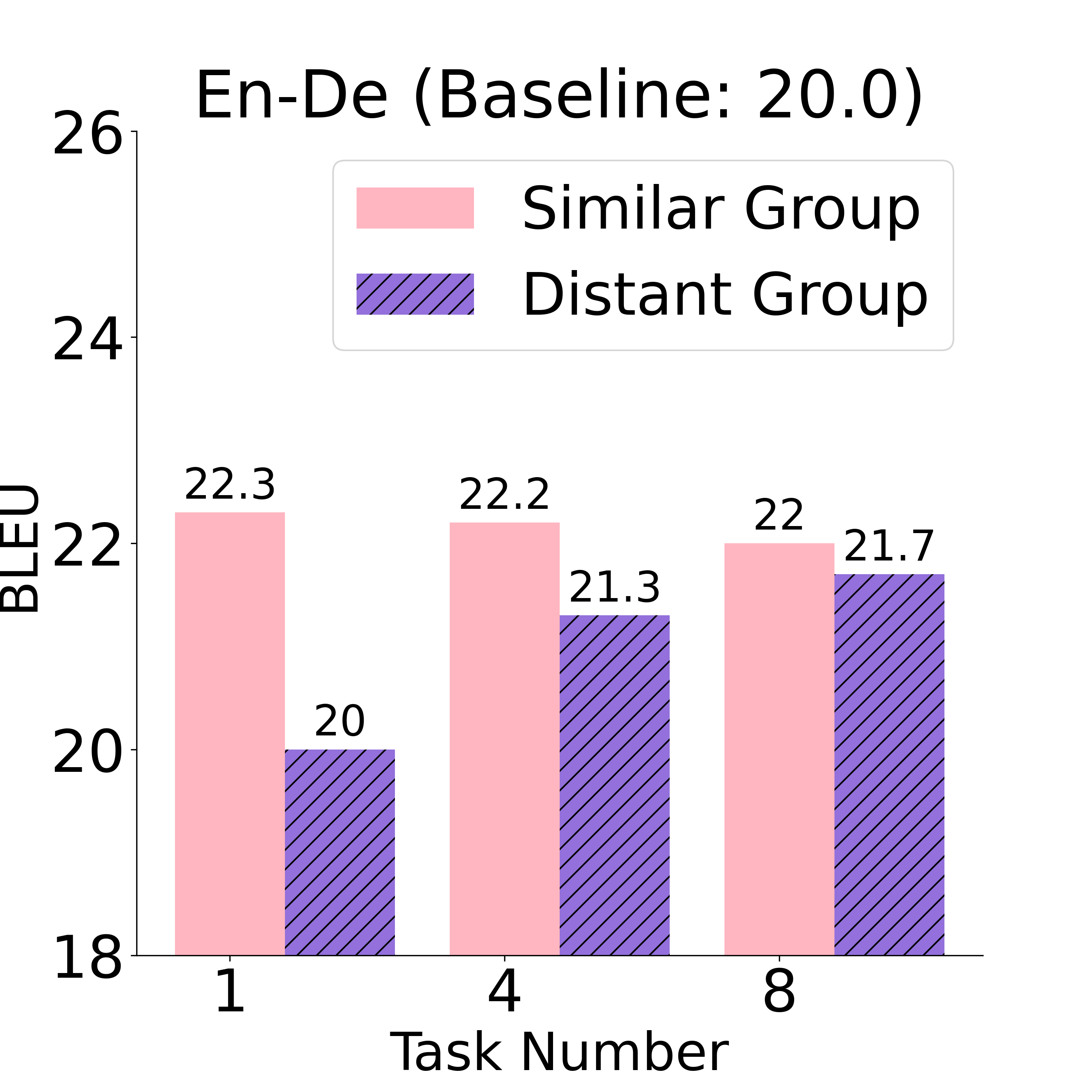}
        \subcaption{Data size: 200\%}
    \end{subfigure}
    \begin{subfigure}{0.23\textwidth}
        \includegraphics[width=\textwidth]{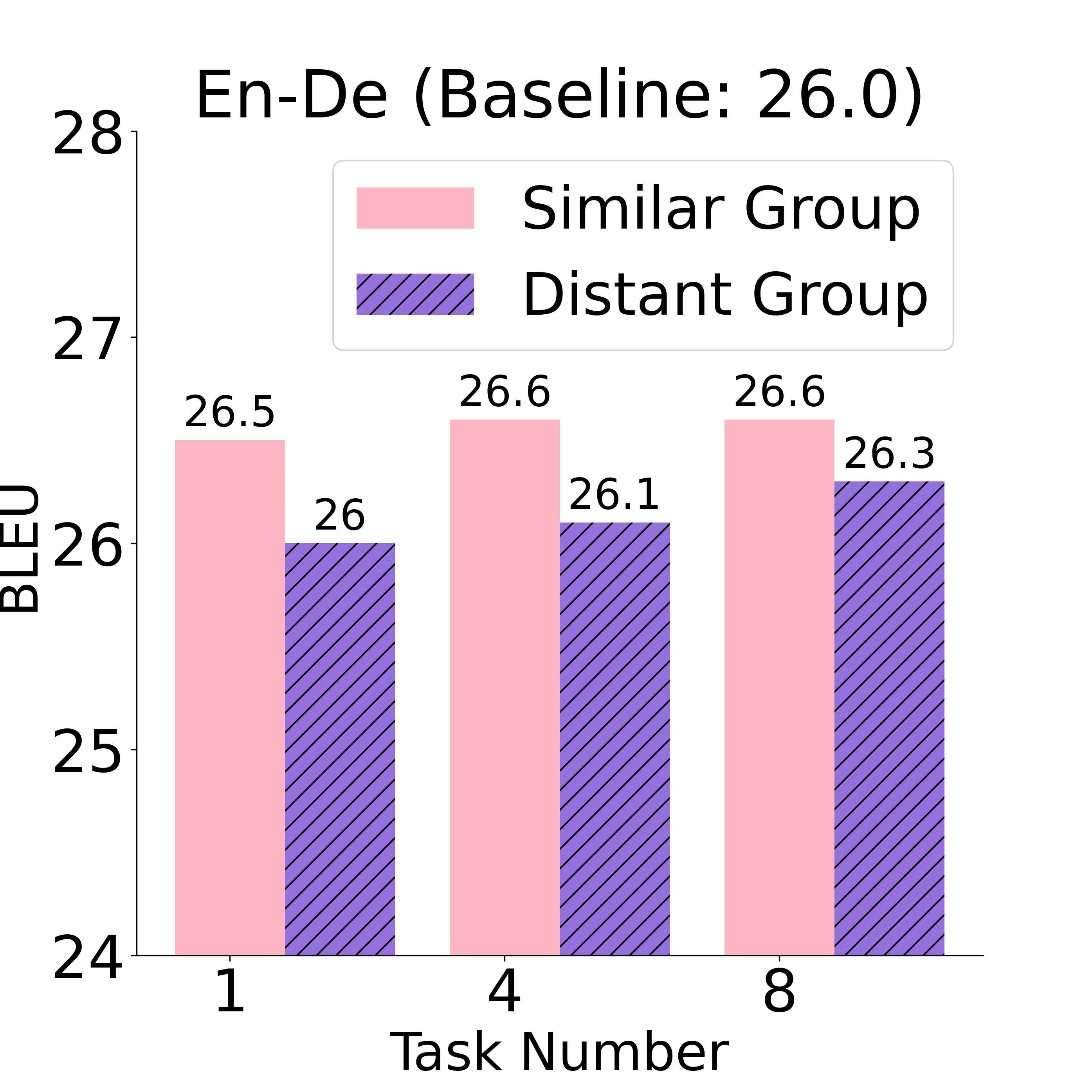}
        \subcaption{Data size: 10\%}
    \end{subfigure}
    \begin{subfigure}{0.23\textwidth}
        \includegraphics[width=\textwidth]{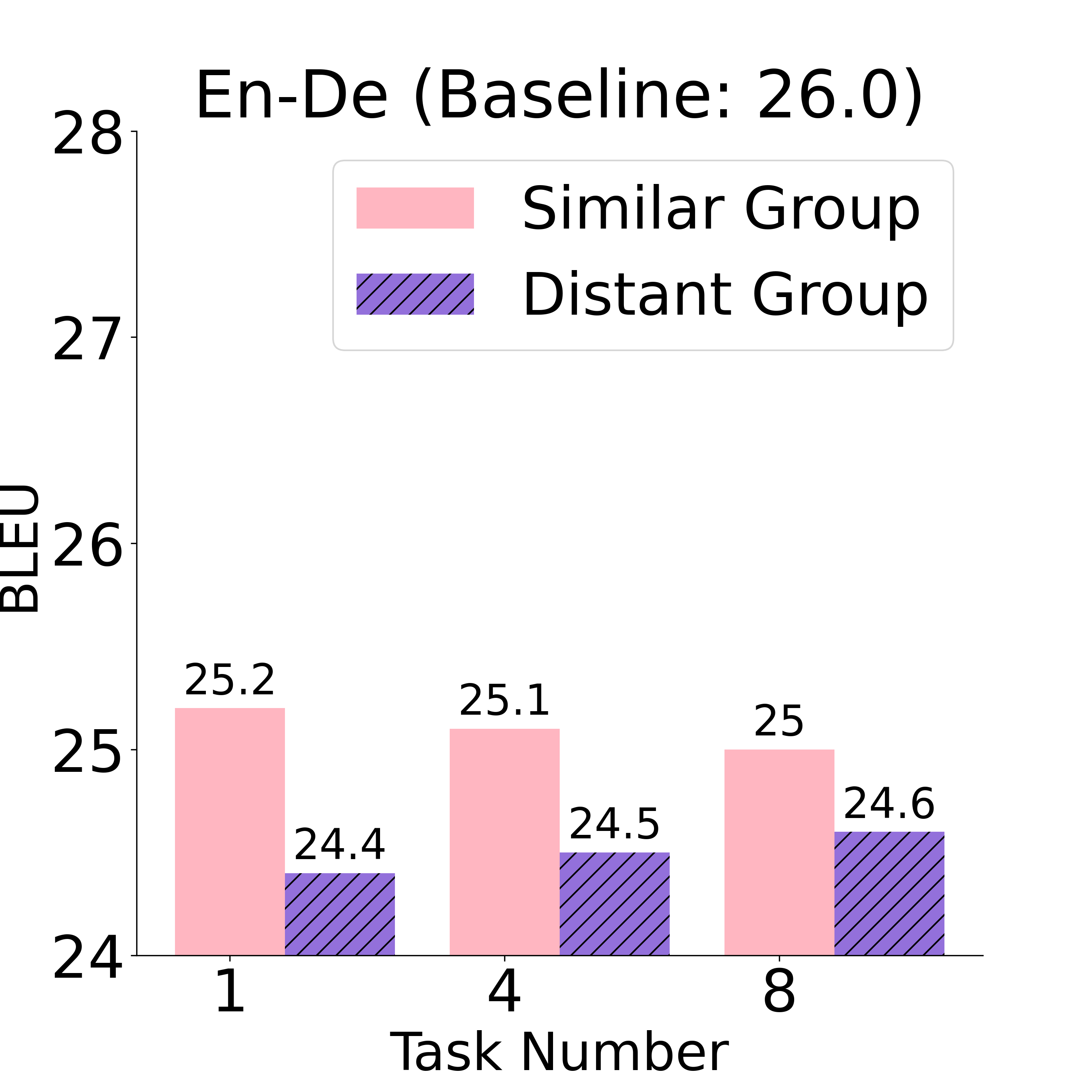}
        \subcaption{Data size: 200\%}
    \end{subfigure}
   
    \caption{Translation quality for En$\rightarrow$De for a low-resource 100K (above), medium-resource 1M (middle) and high-resource 4.5M (below) language pair when training with different auxiliary task numbers and different linguistic groups. Data size represents the total amount of auxiliary target training data.}
    \label{fig:tasknum-high}
    
\end{figure}

We train the main translation task En$\rightarrow$De for different resource levels with an increasing number of auxiliary target language pairs from two groups (Table~\ref{tab:tasknum} in Appendix \ref{appendix:lc}): (1) Similar group: the Germanic\footnote{Due to data scarcity, we pick two target languages from the Romance language family, Galician, and Spanish. Romance and Germanic language families are close.} language family with Latin scripts; 
(2) Distant group: the Slavic language family with Cyrillic scripts. 
The number of target language pairs is set as 1, 4, 8. 
The auxiliary target data size is evenly distributed among all target languages and controlled at 50\% and 1000\% for low-resource, and 10\% and 200\% for medium- and high-resource.
Figure~\ref{fig:tasknum-high} shows the impact of task number when training with auxiliary tasks from different linguistic groups.

\subsubsection{Discussion}

We show that increasing the task number has little impact on target-side knowledge transfer, since our findings are similar for two tasks, see Section~\ref{sec:4.1}: 
(1) Positive transfer highly correlates with linguistic similarity when the auxiliary data size is large; 
(2) small distant auxiliary target data can also benefit the low- and medium-resource main tasks, which is attributed to regularization. 
Interestingly, for the medium-resource settings, increasing the auxiliary target task number from the large-size distant linguistic group (200\%) can mitigate negative transfer to some extent.  
One possible explanation for this is that the negative training signal from one distant language pair becomes weaker when increasing the task number in controlled data size setting. 
This result also corroborates similar findings, where \citet{mitigating} find more than one unrelated language helps the translation task with less data.

In summary, Section~\ref{sec:4} shows how target-side transfer occurs in one-to-many translation. 
Based on the empirical findings on main language pairs, we show that target-side transfer can transfer positive knowledge. Linguistic similarity and target data size mutually play a role in it. 
Meanwhile, we show that the increase in source data cannot be the sole reason for improving one-to-many translation due to the close correlation between translation performance and target data. 
Furthermore, we find that a small amount of distant auxiliary target languages can also improve translation performance. 
These gains cannot be fully attributed to target-side transfer, and we indicate another important factor, i.e., regularization, which is discussed in the next section.

\section{Language Regularization} \label{sec:5}
%The previous section shows the role of knowledge transfer in one-to-many translation and also reveals the minimal impact of increasing the source data. 
The previous section shows low- and medium-resource translation tasks benefit from language regularization. 
In this section, we aim to further investigate the effectiveness of language regularization in one-to-many MT from two angles: generalization ability (Section \ref{sec:ge}) and model calibration (Section \ref{sec:5.2}).
In the end, we provide a simple but effective way to improve machine translation performance with the help of language regularization (Section \ref{sec:5.3}).

\begin{figure}[!t]
\centering
\begin{subfigure}{\linewidth}
    \includegraphics[width =\textwidth]{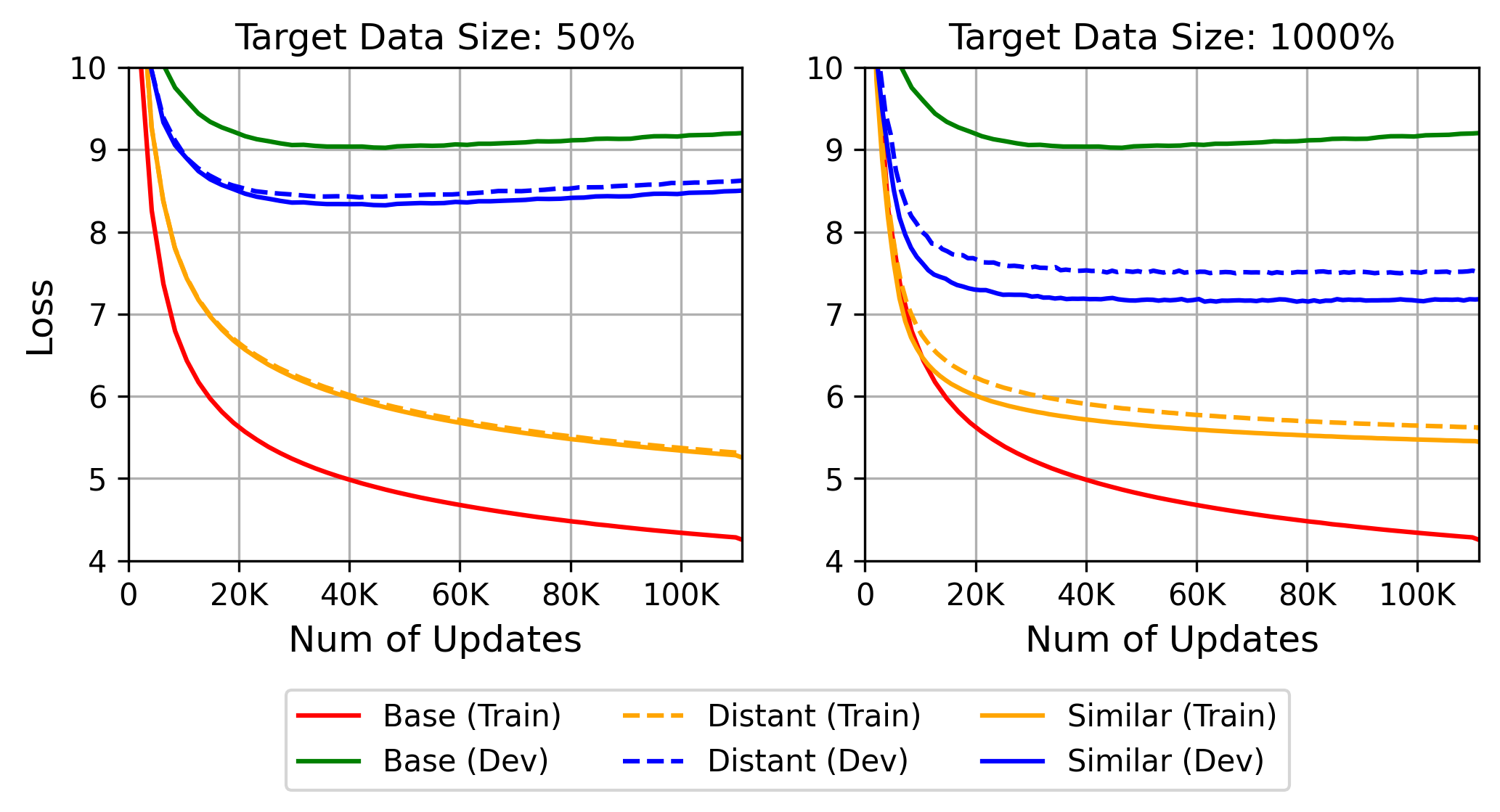}
    \caption{ En$\rightarrow$De in Low-resource (100K) }
    \label{fig:la}
\end{subfigure}
    \begin{subfigure}{\linewidth}
    \includegraphics[width = \textwidth]{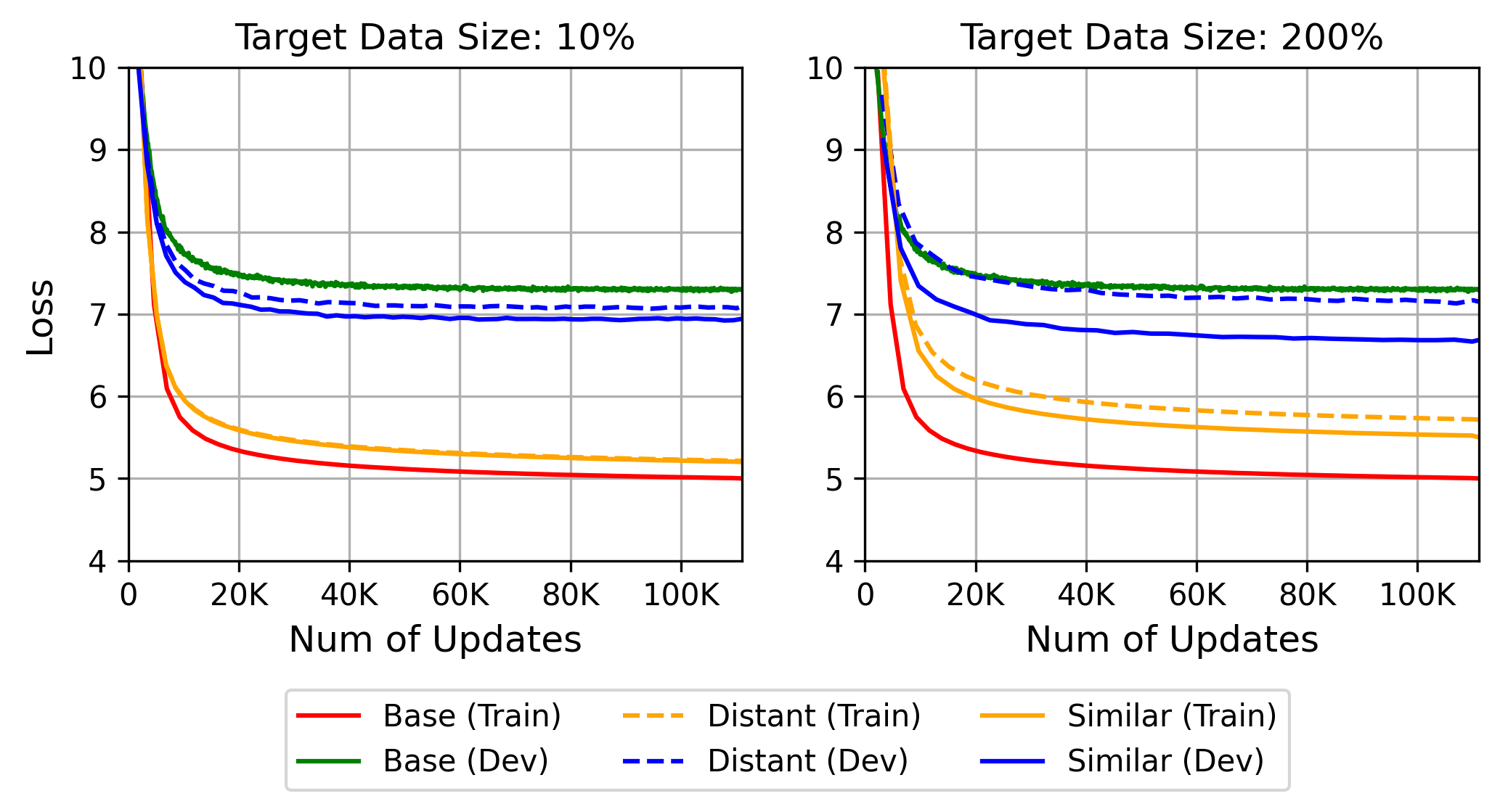}
    \caption{En$\rightarrow$De in Medium-resource (1M) }
    \label{fig:lb}
\end{subfigure}
\caption{Loss curves for En$\rightarrow$De translation tasks under low-resource 100K (a) and medium-resource 1M settings (b), with varying target linguistic groups (similar and distant) and varying auxiliary target data sizes. }
\label{fig:loss}

\end{figure}

\subsection{Reducing Generalization Errors}\label{sec:ge}

Reducing generalization errors is one of the benefits of regularization, which can be reflected by measuring the inconsistency between training and validation performance. 
Here, we show the regularization effects for one-to-many translation by comparing their learning curves for the training and validation losses. 

\subsubsection{Setup}\label{setup} 
Different target languages have various levels of regularization effects. 
We vary the target data linguistic similarity and data size to investigate its impact on generalization ability. 
As we have shown in Section~\ref{sec:4.1}, low- and medium-resource main language pairs benefit from regularization. 
Thus, we choose the multilingual models trained on low- and medium-resource En$\rightarrow$De tasks with two linguistic groups shown in Section~\ref{sec:4.2}. 
%: similar\footnote{Germanic language family and Latin written script} and distant\footnote{Slavic language family and Cyrillic written script}. 
For the low-resource En$\rightarrow$De setting (100K), we select the auxiliary target data size to be 50\% and 1000\% of the low-resource size. 
For the medium-resource En$\rightarrow$De setting (1M), we select the target data size to be 10\% and 200\% of the medium-resource size.
Figure~\ref{fig:loss} shows the learning curves En$\rightarrow$De under different multilingual training settings.

\subsubsection{Discussion}

First, regularization induced by the small size of auxiliary target tasks can reduce the generalization errors in one-to-many translation.
Figure~\ref{fig:la} shows that the baseline bilingual low-resource En$\rightarrow$De model has a large gap between training and validation loss during training. 
This indicates that low-resource models can easily overfit and cannot generalize well to unseen data. 
When training with other target data, the generalization ability for the En$\rightarrow$De task is improved at different levels. 
Surprisingly, 50\% of distant auxiliary data can reduce the validation loss for the main low-resource En$\rightarrow$De task. 
This observation aligns with our finding in Section~\ref{sec:4.2} that distant auxiliary target languages benefit the main task performance. 
It confirms our hypothesis that regularization plays a crucial role by improving generalization ability. 

Second, regularization effects from the large size of auxiliary target tasks can only reduce generalization errors for low-resource language pairs. 
Increasing the auxiliary target data size (+1000\%) leads to better generalization ability for low-resource En$\rightarrow$De, and the linguistically similar group shows slightly better effectiveness than the distant ones. 
This difference shows that positive target-side transfer also helps for better generalization ability since they exhibit a strong and transferable training signal for the main low-resource task. 
The same holds for the medium-resource En$\rightarrow$De setting, see Figure~\ref{fig:lb}. 
However, when training with a large target data size (+200\%), a distant linguistic group cannot further reduce generalization errors. 
This reflects that the role of regularization is not always positive, heavily depending on the target linguistic similarity level and the data size.

\begin{figure}[!t]
    \centering
    \begin{subfigure}{0.48\textwidth}
         \includegraphics[width=\textwidth]{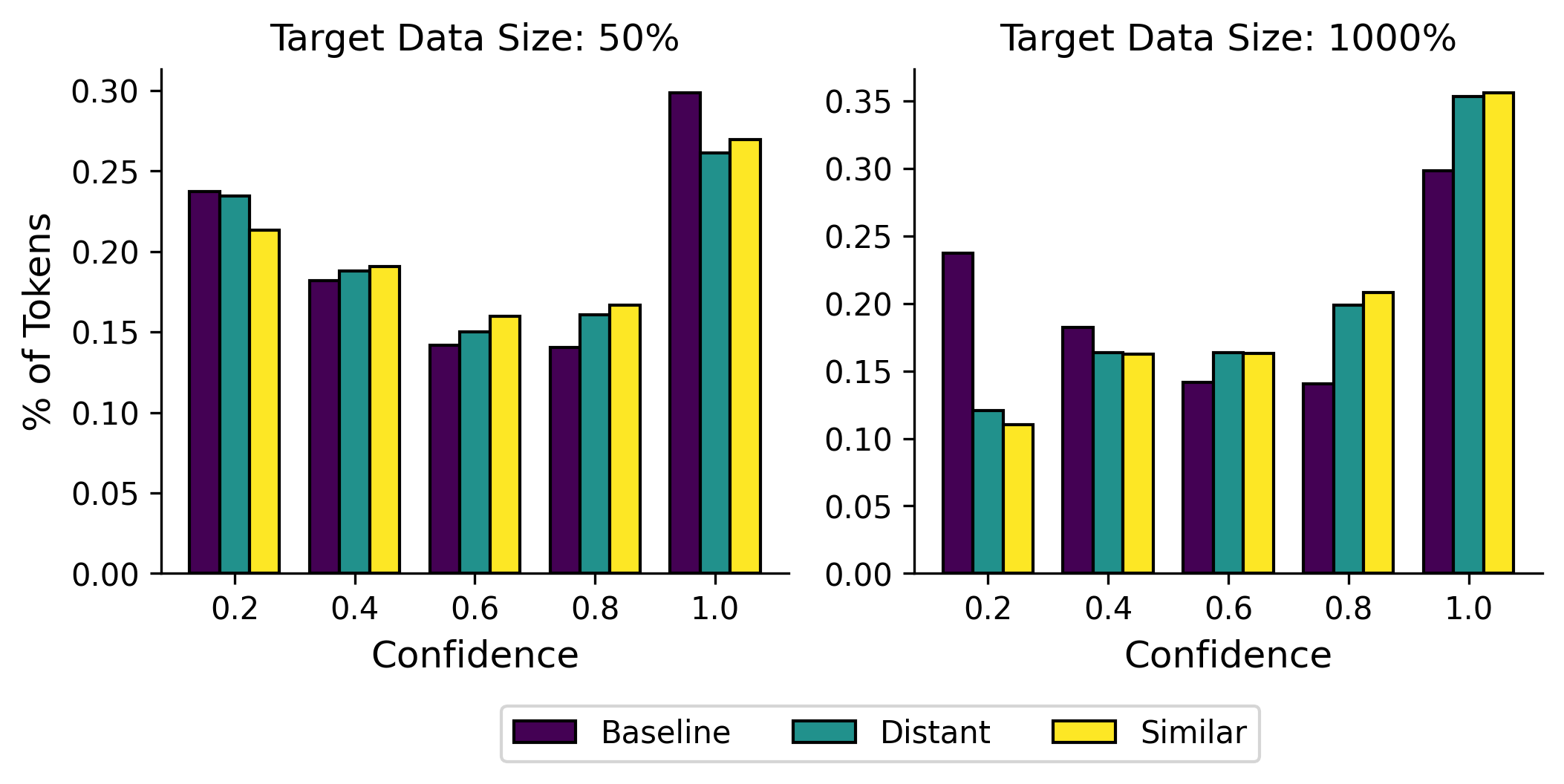}
    \caption{En$\rightarrow$De in Low-resource (100K)}
    
    \label{fig:conf1}
    \end{subfigure}
    \begin{subfigure}{0.48\textwidth}
         \includegraphics[width=\textwidth]{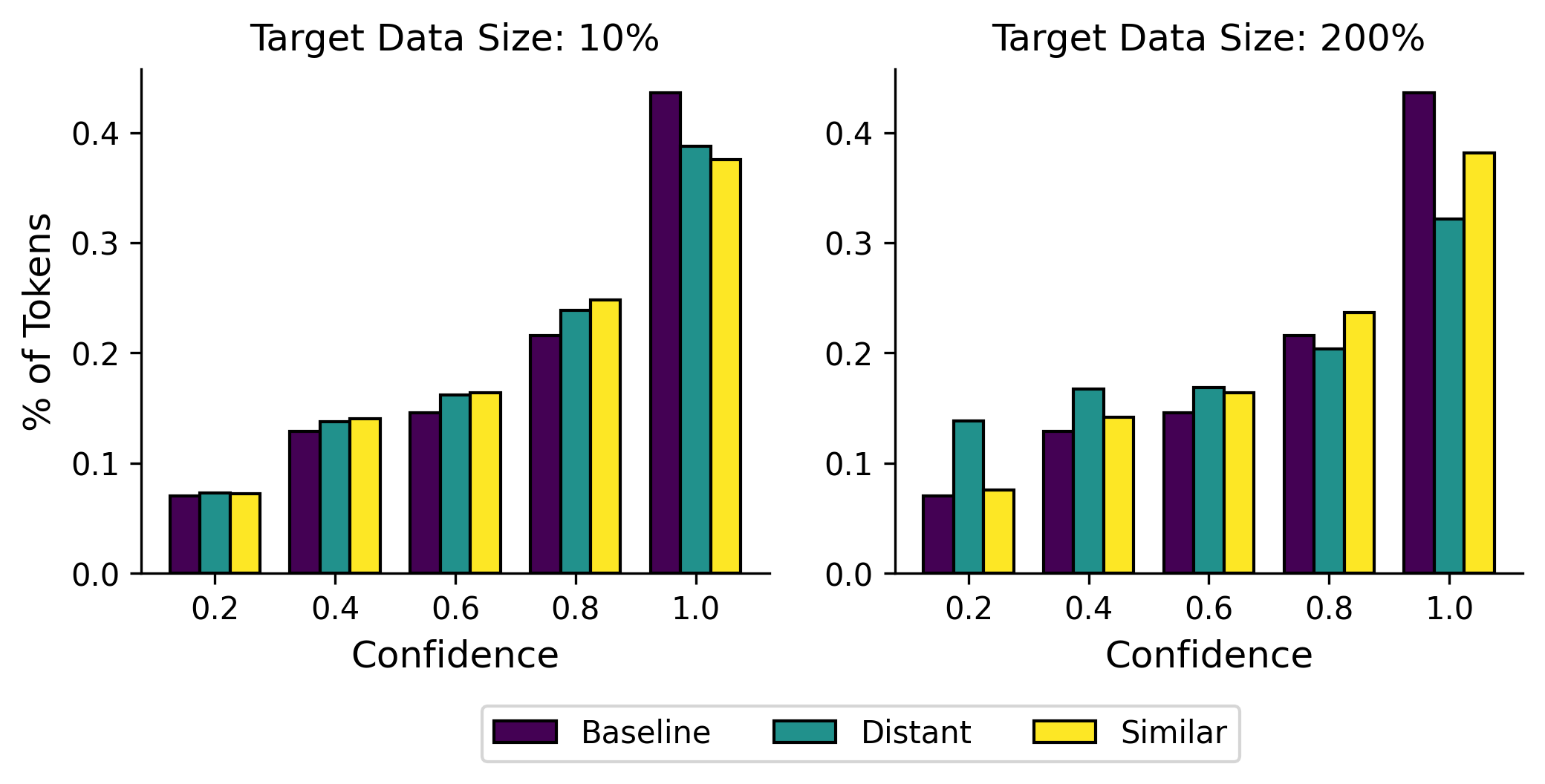}
    \caption{En$\rightarrow$De in Medium-resource (1M)}
    
    \label{fig:conf2}
    \end{subfigure}
    \caption{Confidence histograms for En$\rightarrow$De translation tasks under low-resource (100K) (a) and mid-resource (1M) settings (b), with varying target linguistic groups (similar and distant) and total target data sizes.}
    \label{fig:conf}

\end{figure}

\begin{figure*}[!t]
    \centering
    \begin{subfigure}{0.18\textwidth}
        \includegraphics[width=\textwidth]{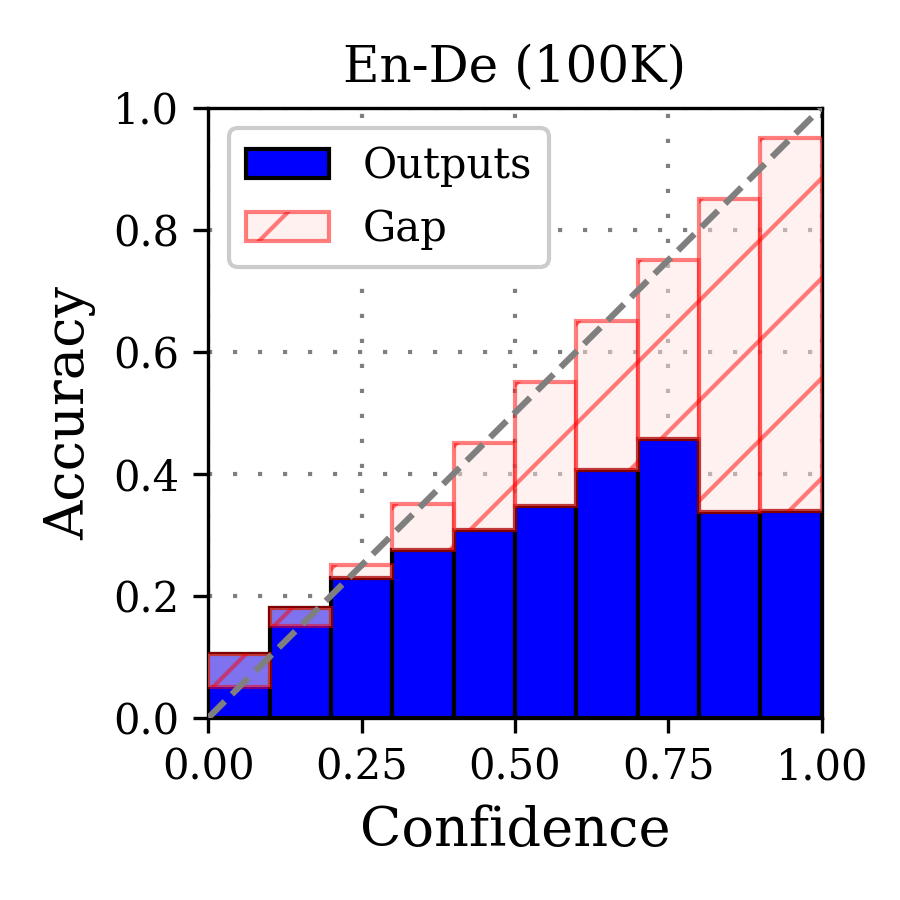}
        \subcaption{InfECE=22.6}\label{fig:ece.base}
    \end{subfigure}
    \begin{subfigure}{0.18\textwidth}
        \includegraphics[width=\textwidth]{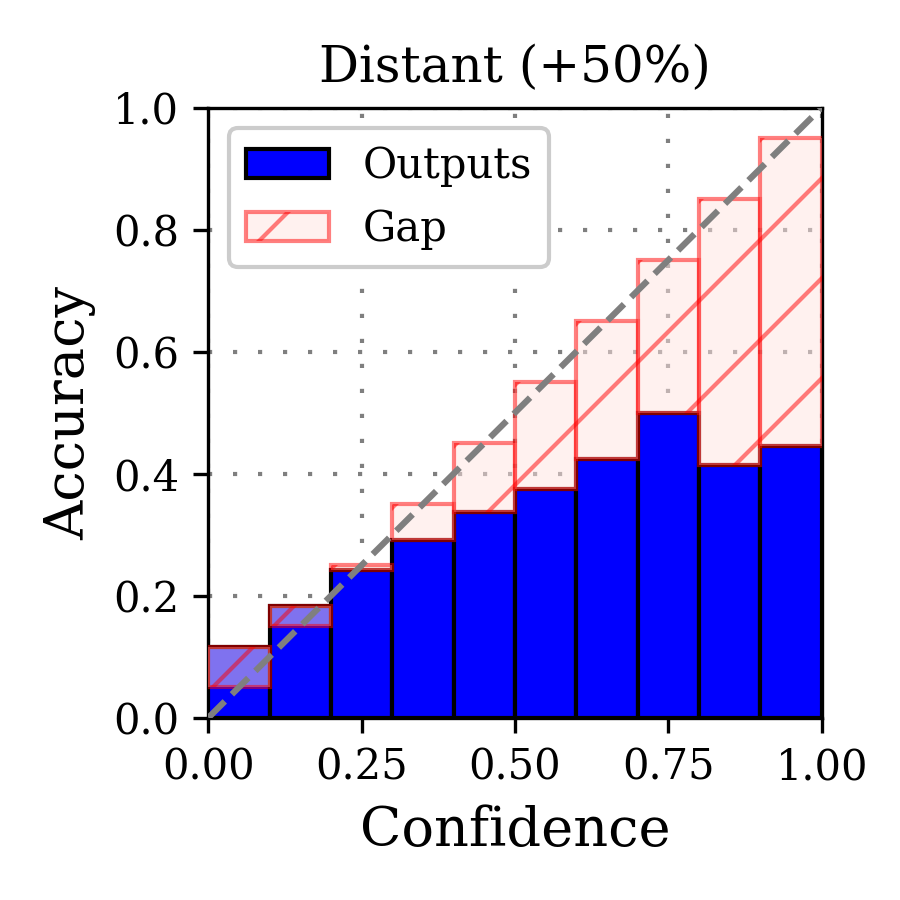}
        \subcaption{InfECE=19.7}
    \end{subfigure}
    \begin{subfigure}{0.18\textwidth}
        \includegraphics[width=\textwidth]{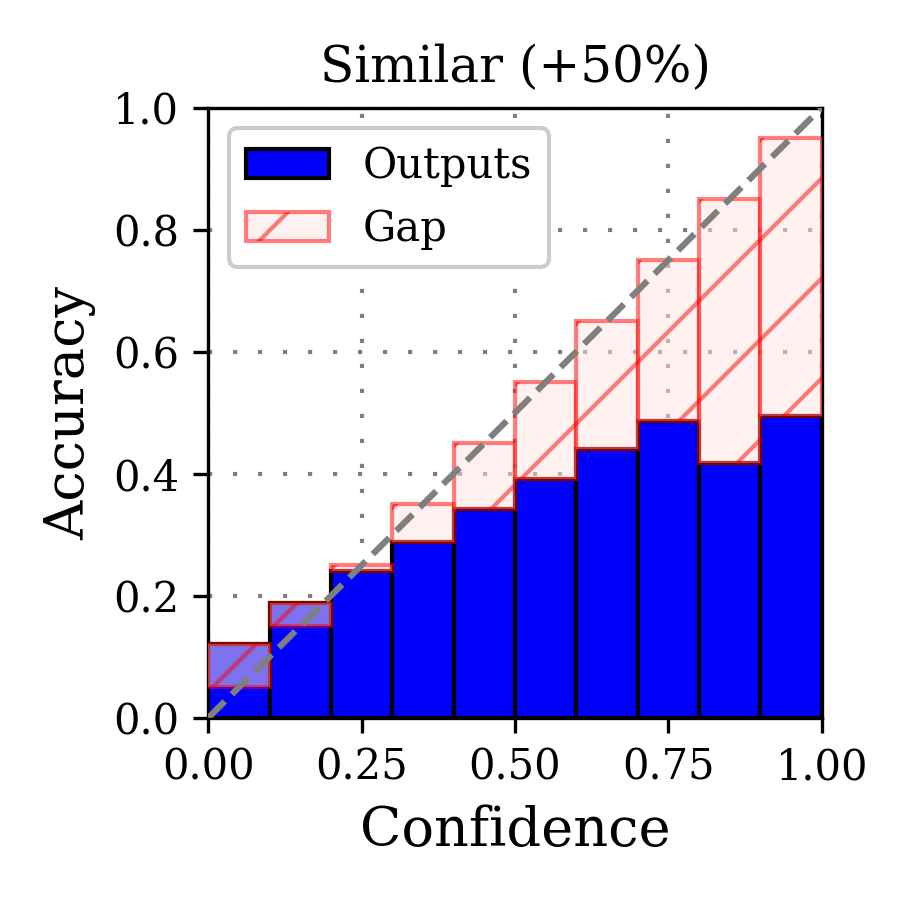}
          \subcaption{InfECE=19.4}
    \end{subfigure}
    \begin{subfigure}{0.18\textwidth}
        \includegraphics[width=\textwidth]{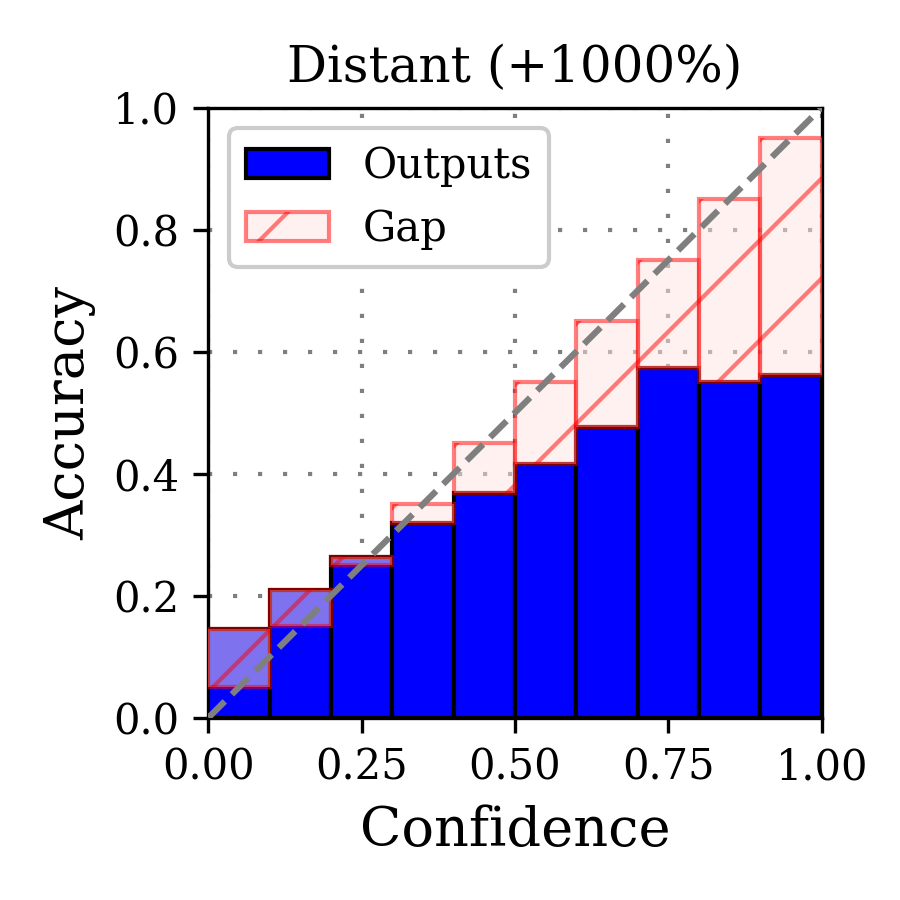}
        \subcaption{ InfECE=18.1}
        \end{subfigure}
    \begin{subfigure}{0.18\textwidth}
        \includegraphics[width=\textwidth]{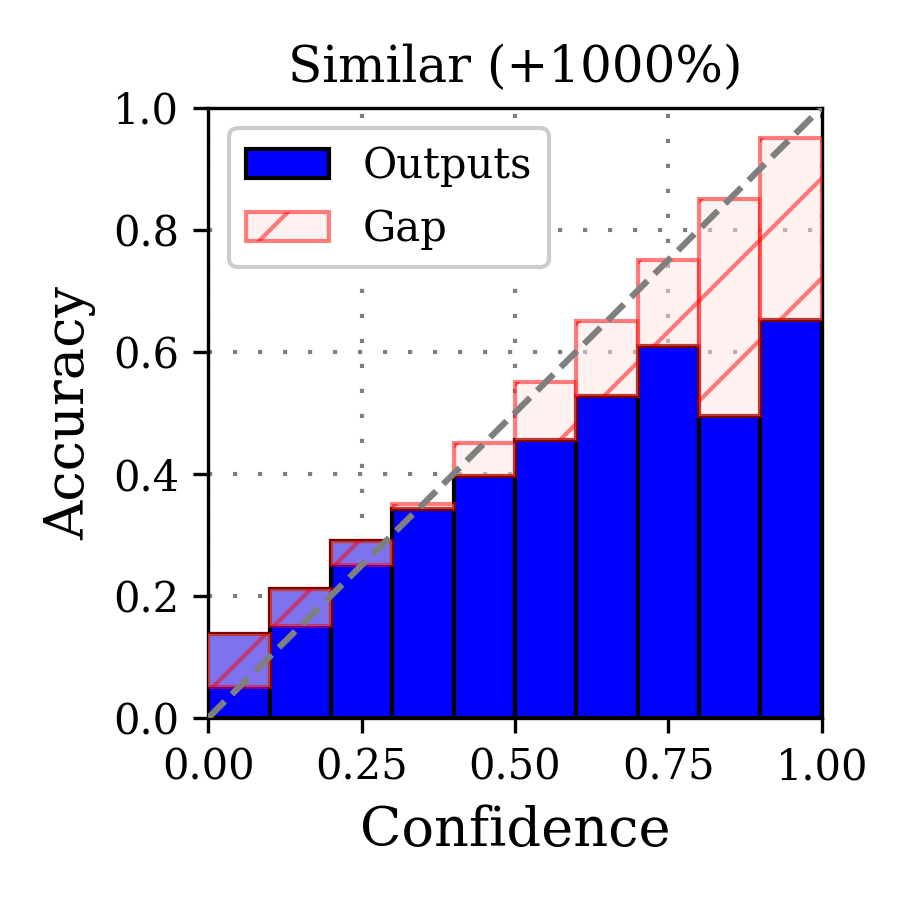}
        \subcaption{InfECE=16.5}
    \end{subfigure}
    \begin{subfigure}{0.18\textwidth}
        \includegraphics[width=\textwidth]{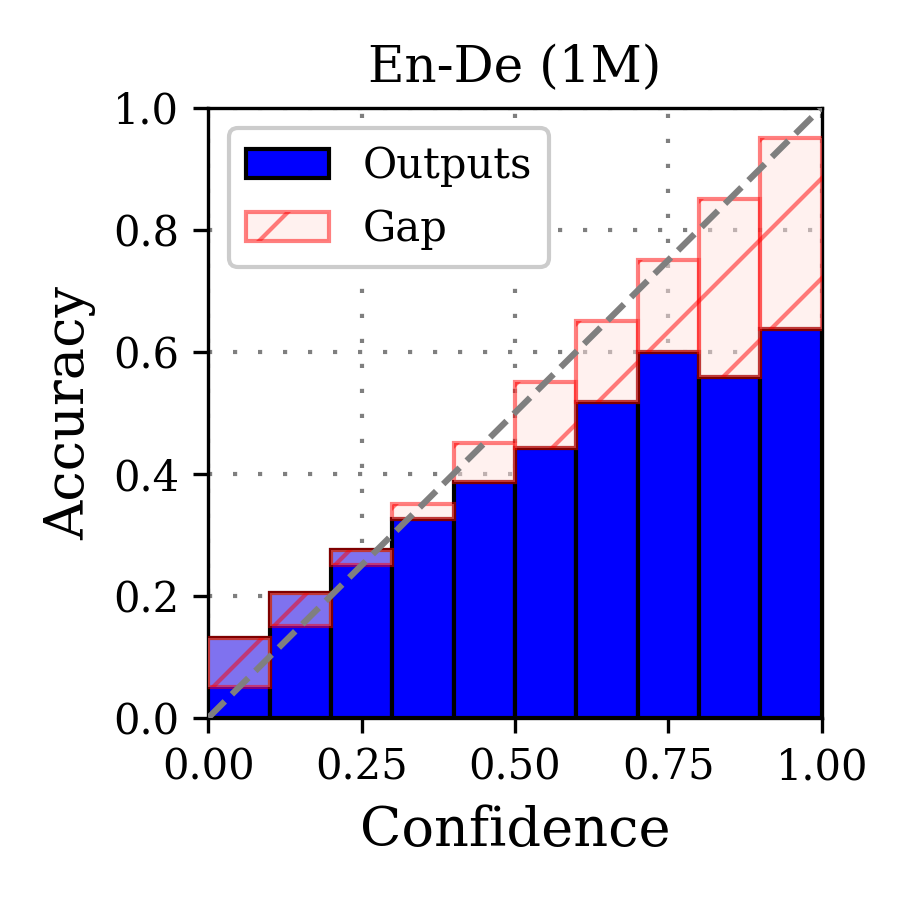}
        \subcaption{InfECE=15.6}
    \end{subfigure}
    \begin{subfigure}{0.18\textwidth}
        \includegraphics[width=\textwidth]{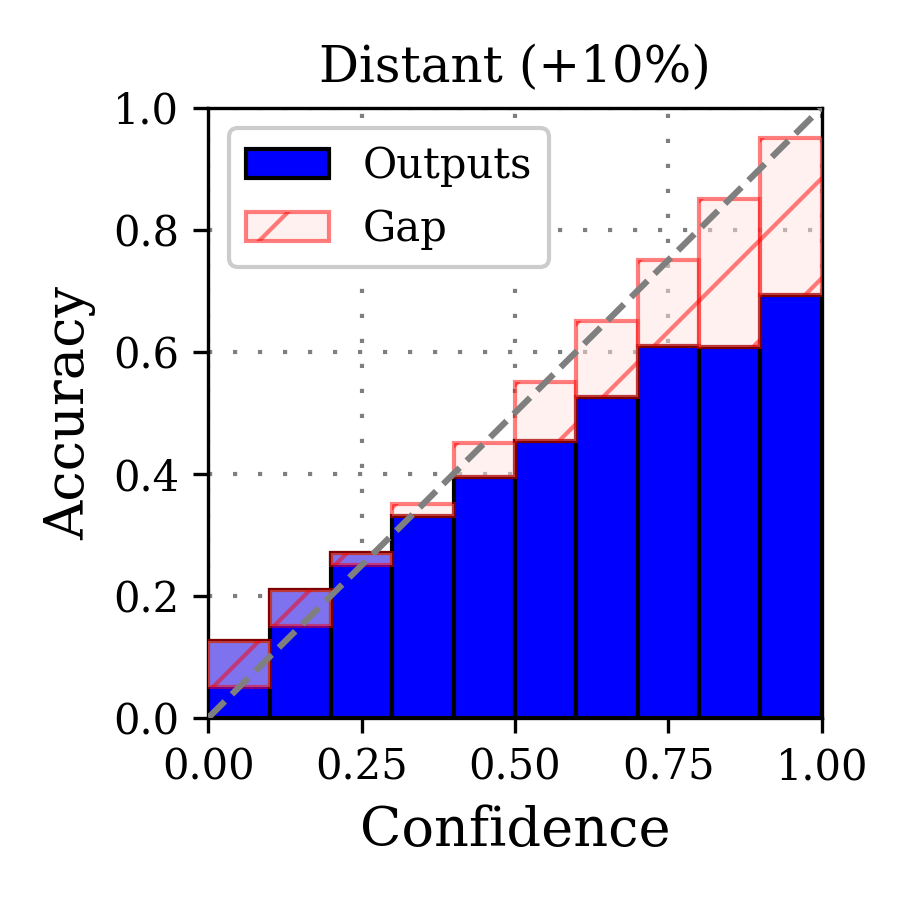}
        \subcaption{InfECE=14.5}
    \end{subfigure}
    \begin{subfigure}{0.18\textwidth}
        \includegraphics[width=\textwidth]{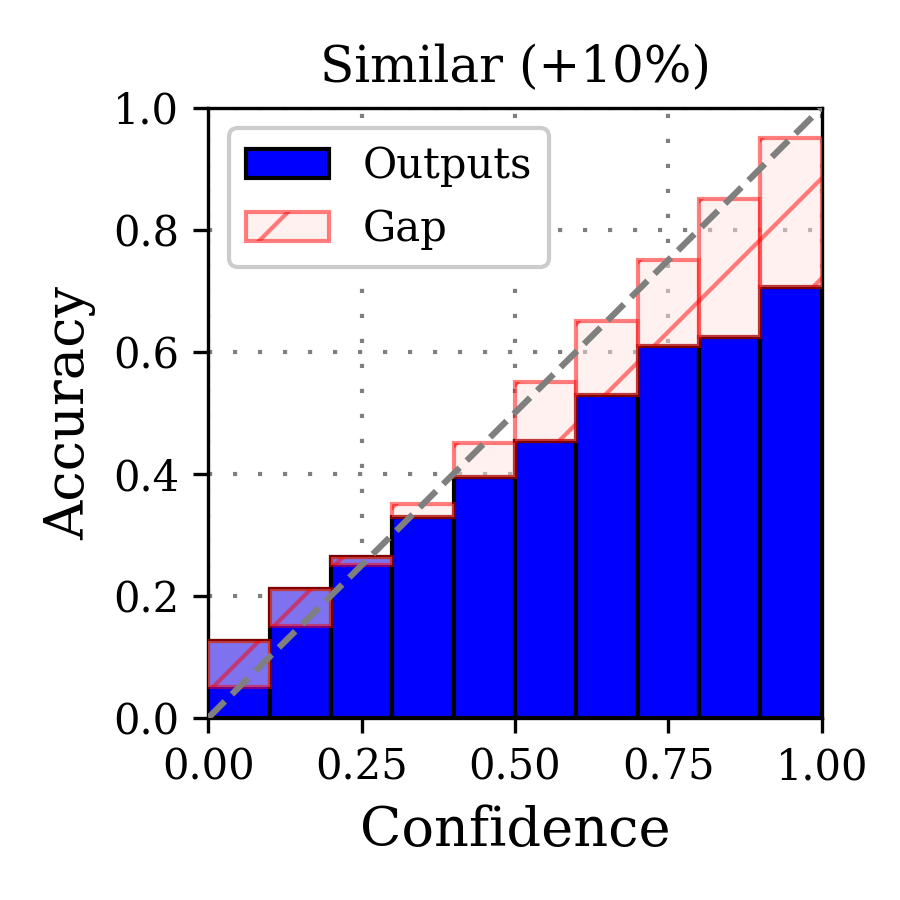}
          \subcaption{InfECE=14.3}
    \end{subfigure}
    \begin{subfigure}{0.18\textwidth}
        \includegraphics[width=\textwidth]{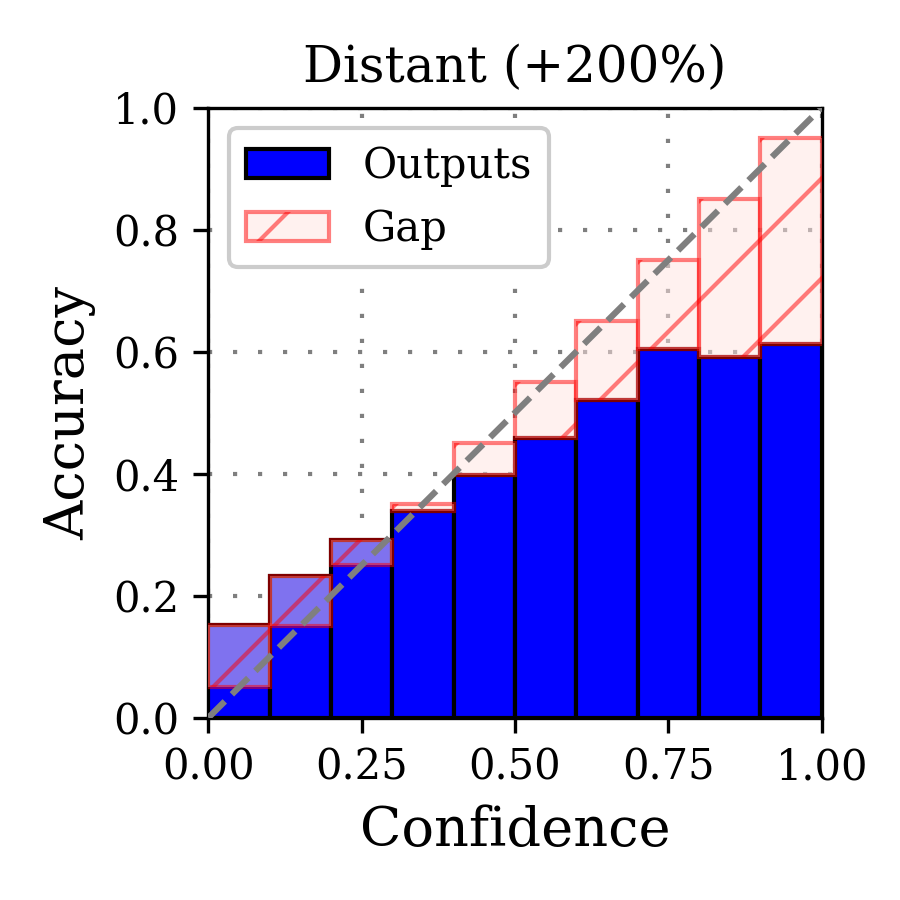}
        \subcaption{ InfECE=17.0}
        \end{subfigure}
    \begin{subfigure}{0.18\textwidth}
        \includegraphics[width=\textwidth]{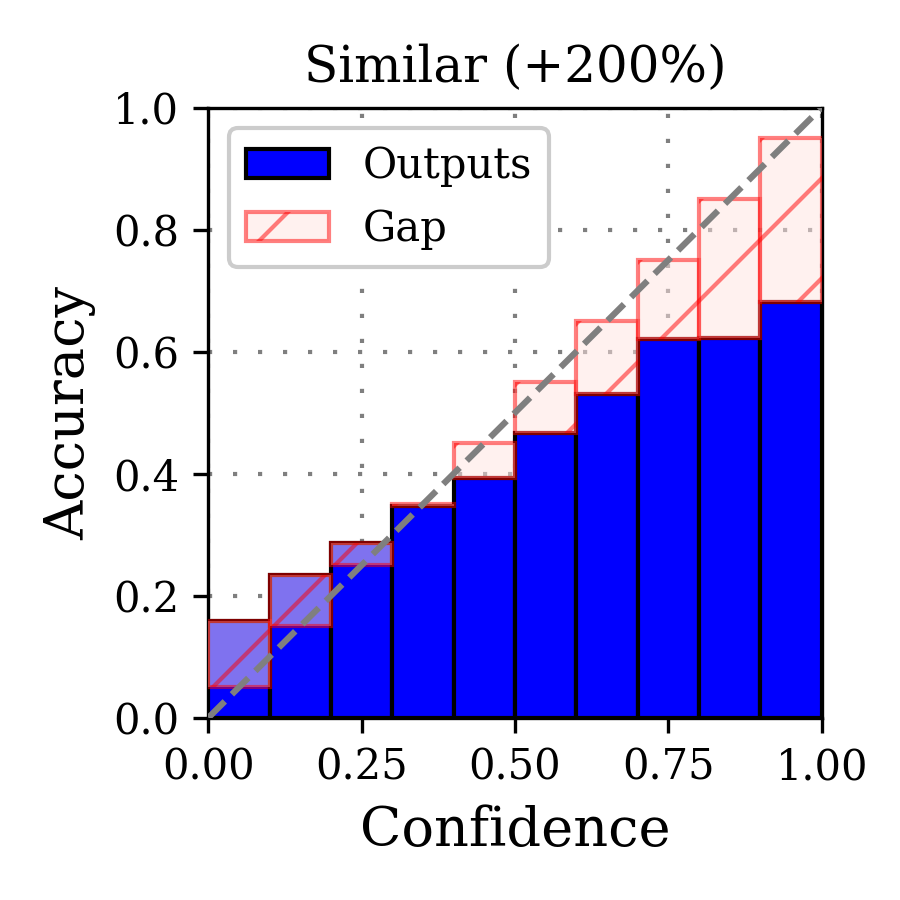}
        \subcaption{InfECE=14.7}
    \end{subfigure}

    \caption{ Reliability diagrams with inference calibration errors (InfECE) on the En$\rightarrow$De test set in the low-resource (above) and medium-resource setting (below). }
    \label{fig:reliable}
 
\end{figure*}

\subsection{Improving Inference Calibration}\label{sec:5.2}
Another benefit of regularization is to increase the model's uncertainty by penalizing output confidence, e.g., label smoothing. 
This regularization technique improves model calibration by making the confidence of its predictions more accurate for true accuracy~\citep{Mller2019WhenDL}.
\citet{Wang2020OnTI} emphasizes the importance of calibrating confidence during inference for MT and regularization is a key factor. 
Motivated by these findings, we aim to investigate whether regularization induced by different target tasks has a similar impact on both output confidence and inference calibration.

In general, model calibration is measured by the expected calibration error (ECE) which calculates the difference in expectation between confidence and accuracy. 
As shown in Equation~\ref{ece}, ECE divides predictions into $M$ bins $\{B_1, ..., B_M\}$ based on their confidence and calculates a weighted average of the bin's accuracy/confidence difference.\footnote{$N$ is the number of prediction samples and $|B_m|$ is the number of samples in the $m$-th bin }

\begin{equation}
    ECE = \sum_{m=1}^M\frac{|B_m|}{N}|acc(B_m)-conf(B_m)|
\end{equation}\label{ece}

In MT, the prediction target token is $\hat{y} = argmax_{y\in{V}}P(y)$ and the confidence is $P(\hat{y})$.  
The accuracy denotes whether the prediction $\hat{y}$ is correct.
However, calculating the prediction accuracy during inference is challenging because it requires building complex alignments between generated tokens and the ground truth. \citet{Wang2020OnTI} propose using the Translation Error Rate metric~\citep{snover-etal-2006-study} to determine the accuracy by measuring the number of edits to change a model output into the ground truth. 
We use their method to analyze inference calibration.

\subsubsection{Setup}
We examine the impact of regularization effects induced by different target data on the model's output confidence and inference calibration for the main En$\rightarrow$De tasks.
We calculate the output confidence histograms and inference calibration errors for the En$\rightarrow$De test set with the same settings as for the multilingual models in Section~\ref{setup}. 
We plot the output confidence histograms in Figure~\ref{fig:conf} where the $x$-axis represents the output confidence scores and the $y$-axis represents the percentage of the number of tokens with those scores.  
In addition, we plot the reliability diagrams in Figure~\ref{fig:reliable} to visualize the representations of model calibration where the $x$-axis is the average weighted confidence and the $y$-axis is the average weighted accuracy. 

\subsubsection{Discussion}

First, regularization from the small size of auxiliary target tasks improves inference calibration by penalizing output confidence.
For example, the main low-resource En$\rightarrow$De translation task shows an over-confidence issue for its bilingual baseline model, see Figure~\ref{fig:ece.base}. 
The model seriously suffers from miscalibration, where the average gaps between confidence and accuracy are large (confidence $>$ accuracy). 
Training with different target tasks could alleviate this issue at various levels. 
The small size of distant auxiliary target tasks can lead to better inference calibration. 
This regularization effect is achieved by penalizing over-confidence output ($>0.9$) to enhance the model inference calibration, as shown in Figure~\ref{fig:conf1}. 
These findings also align well with the medium-resource setting (1M). 
The relatively small size of auxiliary target tasks (10\%) benefits inference calibration from penalizing over-confident output, as shown in Figure~\ref{fig:conf2}. 

Second, regularization in the large-size auxiliary target tasks improves inference calibration by improving translation accuracy.
Unlike in the small data (50\%) scenario, which penalizes over-confident output probabilities to benefit the task, training with a large size of auxiliary target language pairs mainly helps the low-resource En$\rightarrow$De task improve translation accuracy to benefit inference calibration.  
Since similar language pairs share similar lexical and word order knowledge with the low-resource En$\rightarrow$De task, they improve accuracy more effectively.

\begin{table}[!t]
    \centering
     \resizebox{0.85\linewidth}{!}{
    \begin{tabular}{cccc}
    \toprule
        Main Task  & Auxiliary Task  & BLEU & $\triangle$ \\
        \hline 
         \multirow{3}{*}{En$\rightarrow$De }  & En$\rightarrow$De &  28.4 & -0.2  \\  
           & En$\rightarrow$Nl & 28.3 & -0.3  \\  
            & En$\rightarrow$Zh & 29.0  & +0.6\\  
          \bottomrule
    \end{tabular}}
    \caption{The main task of En$\rightarrow$De BLEU scores with using larger model by adding 10\% auxiliary tasks; $\triangle$ represents the BLEU changes with the En$\rightarrow$De baseline.}
    \label{tab:cheap}

\end{table}

\subsection{Regularization Effect in Larger Models}\label{sec:5.3}
Sections~\ref{sec:ge} and \ref{sec:5.2} show that utilizing small distant auxiliary data can prevent overfitting translation models by regularization, particularly for low- and medium-resource language pairs. 
To further verify the impact of language regularization at a larger scale, we increase the model size from Transformer-Base (93M parameters) to Big (274M parameters) and utilize 10\% of different auxiliary data to train with a high-resource En$\rightarrow$De (4.5M) translation task\footnote{https://www.statmt.org/wmt14}. 
Table~\ref{tab:cheap} shows that 10\% of distant auxiliary data En$\rightarrow$Zh can help  improve the bilingual baseline while adding the same target languages or similar ones cannot. 
This finding further shows the effectiveness of language regularization for optimizing machine translation performance.

\section{Conclusion}

In this work, we disentangle the roles of knowledge transfer and language regularization in one-to-many MMT. 
In contrast to previously held assumptions, we show that target-side knowledge transfer \textit{does} play an important role in one-to-many MMT, influenced by several dominant factors: auxiliary target data size, linguistic similarity, and the number of auxiliary target tasks.
This finding also shows that the increased amount of source data \textit{does not} explain all transfer.
Future work can leverage this information to encourage different language pairs to have similar word representations to achieve maximum positive transfer. 
Surprisingly, we find that using a small amount of linguistically distant auxiliary target data acts as an effective regularizer resulting in translation performance gains. 
%This finding provides a simple but effective way to boost the performance of real-world low- and medium-resource language pairs, especially when similar target languages are not available.  
This form of language regularization shows its effectiveness by benefiting generalization ability and inference calibration. 
Our findings on language regularization provide a simple but effective way to boost the translation performance of real-world low- and medium-resource language pairs, especially when similar target languages are not available. 
Future work can further explore the optimization of multilingual training by leveraging distant auxiliary data.

\section*{Limitations}
We acknowledge several limitations in our work. 
To directly understand the impact of knowledge transfer, source data, and regularization in one-to-many translation, we only observe the performance changes for one selected main language pair. 
Although translation results for auxiliary language pairs are provided in Appendix~\ref{addi-results}, further analysis of the dynamic performance trade-off between main and auxiliary language pairs is worthwhile to explore. 
Another limitation of our work is about the MMT setting, where we only work in one-to-many MT, while future work should extend it to many-to-many settings and explore the impact of adding multiple source languages.

\section*{Acknowledgments}

This research was funded in part by the Netherlands Organization for Scientific Research (NWO) under project number VI.C.192.080. We would like to thank Di Wu, Shaomu Tan, David Stap, Wafaa Mohammed and Baohao Liao for their useful feedback and discussion. We would also like to thank the reviewers for their feedback. Yan Meng also thanks Yuheng Huo for encouragement and inspiration. 

\bibliography{anthology,custom}

\clearpage

\appendix

\section{Language Choices}
\label{appendix:lc}

Table~\ref{tab:tasknum} shows two linguistic groups trained with the main language pair.

\begin{table}[htbp]
\centering
    \begin{subtable}{0.48\linewidth}
    
    \resizebox{\linewidth}{!}{
        \begin{tabular}{cc|cc}
        \toprule
           ISO & Lang. & Family & Script \\
              \midrule
           % \textbf{De} & \textbf{German} &  \textbf{Germanic} & \textbf{Latin} \\
           % \hdashline
            Af & Afrikaans & Germanic & Latin \\
             Da  & Danish  & Germanic & Latin \\
             Nl  & Dutch  & Germanic & Latin \\
            Is  & Icelandic & Germanic & Latin \\
             No  & Norwegian  & Germanic & Latin \\
             
            Sv  & Swedish  & Germanic & Latin \\
            Gl  & Galician  & Romance & Latin \\
            Es  & Spanish  & Romance & Latin \\
            \bottomrule
        \end{tabular}}
    \end{subtable}
    \hfill
        \begin{subtable}{0.48\linewidth}
        
    \resizebox{\linewidth}{!}{
        \begin{tabular}{cc|cc}
        \toprule
            ISO & Lang. & Family & Script \\
              \midrule
          %  \textbf{Ru} &\textbf{Russian} & \textbf{Slavic} &\textbf{ Cyrillic} \\
          % \hdashline
            Bg &  Bulgarian &Slavic & Cyrillic \\
            Cs &  Czech &Slavic & Cyrillic \\
            Mk &  Macedonian &Slavic & Cyrillic \\
            Pl &  Polish &Slavic & Cyrillic \\
            Sr &  Serbian &Slavic & Cyrillic \\
            Sk &  Slovak &Slavic & Cyrillic \\
            Sl &  Slovenian &Slavic & Cyrillic \\
            Uk &  Ukrainia &Slavic & Cyrillic \\

            \bottomrule
        \end{tabular}}

    \end{subtable}
    
    \caption{Two groups of auxiliary target languages.}  %\textbf{Bold} designates the main target languages (De, Ru). }
    \label{tab:tasknum}
\end{table}

\section{Model Parameters}
\label{sec:appendix-model}

We follow the setup of the Transformer-base and Transformer-big models \cite {NIPS2017_3f5ee243}. 
For each model, the number of layers in the encoder and in the decoder
is $N = 6$. 
For Transformer-base, we employ $h = 8$ parallel attention layers or heads. 
The dimensionality of input and output is $d_model = 512$, and the inner layer of feed-forward networks has dimensionality $d_{ff} = 2048$. 
For Transformer-big, we employ $h = 16$ parallel attention layers or heads. 
The dimensionality of input and output is $d_model = 1024$, and the inner layer of feed-forward networks has dimensionality $d_{ff} = 4096$. 

\section{Dataset Statistics}\label{sec:appendix-data}
The data statistics of mimic and real-world main language pairs are shown in Table~\ref{tab:ds} and Table~\ref{tab:real}. The data statistics of joint training target language pairs are shown in Table~\ref{tab:ds1}. 

\begin{table}[htbp]
    \centering
    \resizebox{0.46\textwidth}{!}{
    \begin{tabular}{ccccc}
    \toprule
       Language   & ISO & Dataset Source & Validation Set & Test Set  \\
       \midrule
        German  &  De &  WMT14   & WMT14   & WMT14 \\
        Spanish & Es & WMT13  & WMT13 & WMT13 \\
        Russian & Ru & WMT22 & WMT22& WMT22  \\
        \bottomrule
    \end{tabular}}
    \caption{The data statistics of main low- and medium-resource language pairs. For each language, we display the ISO code, language name, sampled training dataset source, validation set, and test set. Sampled training low-resource dataset size: 100K, sampled training medium-resource dataset size: 1M.  }
    \label{tab:ds}
\end{table}

\begin{table}[htbp]
    \centering
    \resizebox{0.46\textwidth}{!}{
    \begin{tabular}{ccccc}
    \toprule
       Language   & ISO & Dataset Source & Validation Set & Test Set  \\
       \midrule
        Sinhala  &  Si &  OPUS  & OPUS   & OPUS \\
        Belarusian & Be & OPUS & OPUS & OPUS  \\
        \bottomrule
    \end{tabular}}
    \caption{The data statistics of real-world main low- and medium-resource language pairs. For each language, we display the ISO code, language name, sampled training dataset source, validation set, and test set. Training size for En-Si: 979109, for En-Be: 67312. }
    \label{tab:real}
\end{table}

\begin{table}[htbp]
    \centering
    \resizebox{0.46\textwidth}{!}{
    \begin{tabular}{cccc}
    \toprule
       Language   & ISO & Dataset Source   & Validation/Test Set  \\
       \midrule
         Estonia & Et & CCMatrix & CCMatrix \\
        Chinese & Zh & CCMatrix &  CCMatrix \\
        Portuguese & Pt & CCMatrix  & CCMatrix  \\
        Ukrainian & Uk & CCMatrix & CCMatrix \\
        Czech & Cs &  CCMatrix & CCMatrix \\
        Dutch & Nl & CCMatrix & CCMatrix \\
        Afrikaans & Af & CCMatrix & CCMatrix \\
        Danish & Da & CCMatrix & CCMatrix \\
        Icelandic & Is & CCMatrix & CCMatrix \\
        Norwegian & No & CCMatrix & CCMatrix \\
        Swedish & Sw & CCMatrix & CCMatrix \\
        Galician & Gl & CCMatrix & CCMatrix \\
        Bulgarian & Bg & CCMatrix & CCMatrix \\
        Macedonian & Mk & CCMatrix & CCMatrix \\
        Polish & Pl & CCMatrix & CCMatrix \\
        Serbian & Sr & CCMatrix & CCMatrix \\
        Slovak & Sk & CCMatrix & CCMatrix \\
        Slovenian & Sl & CCMatrix & CCMatrix \\

        \bottomrule 
        
    \end{tabular}}
    \caption{ The data statistics of auxiliary training target language pairs. For each language, we display the ISO code, language name, sampled training dataset source, and validation set. The validation and test sets from CCMatrix, are randomly sampled from the CCMatrix corpus, each containing 2000 samples.   }
    \label{tab:ds1}
\end{table}

\section{Additional Results}\label{addi-results}
Here, we show all auxiliary language BLEU scores in Table \ref{tab:low-resource-aux} and \ref{tab:mid-resource-aux}. 

\begin{table}[!t]

    \centering
    
 \begin{subtable}{\linewidth}
 \centering
   \resizebox{0.8\linewidth}{!}{
    \begin{tabular}{r|rrrr}
    \toprule
        \multicolumn{5}{c}{\textbf{En$\rightarrow$De } }\\
    
    $\alpha\%$ &  en$\rightarrow$nl & en$\rightarrow$et & en$\rightarrow$ru & en$\rightarrow$zh \\
    \midrule
       10\% & 8.9\textsubscript{0.2} & {6.2\textsubscript{0.7}} & {6.0\textsubscript{0.6}} & {5.5\textsubscript{0.5}}  \\
       50\% & {11.9\textsubscript{0.2}} & {11.2\textsubscript{0.3}}  &{10.2\textsubscript{0.3}}  & {9.8\textsubscript{0.3}}  \\
       100\% & {20.3\textsubscript{0.2}} & {11.9\textsubscript{0.4}}  & {13.7\textsubscript{0.2}} &{12.3\textsubscript{0.4}}  \\
       500\% & {23.7\textsubscript{0.3}}  &{14.3\textsubscript{0.1}}& {{17.6}\textsubscript{0.3}}   &  {15.6\textsubscript{0.2}} \\
       1000\%  & {26.4\textsubscript{0.2}} & {15.3\textsubscript{0.5}} & {18.5\textsubscript{0.1}} & {16.7\textsubscript{0.3}} \\
    \bottomrule

    \end{tabular}}
    
       \end{subtable}
\begin{subtable}{\linewidth}
\centering
   \resizebox{0.8\linewidth}{!}{
    \begin{tabular}{r|rrrr}
    \toprule
        \multicolumn{5}{c}{\textbf{En$\rightarrow$Ru  } }\\
    
    $\alpha\%$  & en$\rightarrow$uk & en$\rightarrow$cs & en$\rightarrow$de & en$\rightarrow$zh \\
    \midrule
       10\%   & 8.8\textsubscript{0.6} &  7.6\textsubscript{0.6} & 7.8\textsubscript{0.2} & {5.0}\textsubscript{0.2}   \\
       50\%  & 15.0\textsubscript{0.5} & {12.2}\textsubscript{0.2}  & 10.2\textsubscript{0.3}& 9.3\textsubscript{0.1} \\
       100\% & 18.3\textsubscript{0.3}  & 12.6\textsubscript{0.1}  & 11.0\textsubscript{0.2} & 12.5\textsubscript{0.4} \\
       500\% & 22.7\textsubscript{0.2}& {{14.2}\textsubscript{0.1}}   &  {16.8\textsubscript{0.2}} & 15.1\textsubscript{0.1}\\
       1000\%  & {23.4\textsubscript{0.1}} &{14.7}\textsubscript{0.2} &{18.9\textsubscript{0.2}} & 16.2\textsubscript{0.2} \\
    \bottomrule

    \end{tabular}}
   
       \end{subtable}

        \begin{subtable}{\linewidth}
        \centering
   \resizebox{0.8\linewidth}{!}{
    \begin{tabular}{r|rrrr}
    \toprule
        \multicolumn{5}{c}{\textbf{En$\rightarrow$Es } }\\
    
    $\alpha\%$   & en$\rightarrow$pt & en$\rightarrow$nl & en$\rightarrow$ru & en$\rightarrow$zh \\
    \midrule
       10\% &  {9.2\textsubscript{0.4}} &  {8.6\textsubscript{0.6}} & {6.2\textsubscript{0.3}} & {5.1\textsubscript{0.8}}   \\
       50\%  & {12.3\textsubscript{0.3}}  & {11.3\textsubscript{0.6}}  & {10.0\textsubscript{0.2}} & {{9.2}\textsubscript{0.3}} \\
       100\%  & {20.5\textsubscript{0.3}}  & {15.2\textsubscript{0.3}} & {11.5\textsubscript{0.3}} & {{12.5}\textsubscript{0.2}} \\
       500\% & {23.2\textsubscript{0.2}}& {{18.2}\textsubscript{0.3}}   &  {16.5\textsubscript{0.3}} & {15.6\textsubscript{0.2}}\\
       1000\%  &  {26.2\textsubscript{0.4}} & {19.6\textsubscript{0.1}} & {18.6\textsubscript{0.3}} & {16.4\textsubscript{0.1}} \\
    \bottomrule

    \end{tabular}}
    
       \end{subtable}

    \caption{BLEU scores for the auxiliary language pairs in a low-resource setting (100K) when training with main language pairs: En$\rightarrow$De, En$\rightarrow$Es, and En$\rightarrow$Ru. $\alpha\%=10, 50, 100, 500, 1000$ represents the proportion of the low-resource setting size.  }
    \label{tab:low-resource-aux}

\end{table}

\begin{table}[!t]

    \centering
    
 \begin{subtable}{\linewidth}
 \centering
   \resizebox{0.8\linewidth}{!}{
    \begin{tabular}{r|rrrr}
    \toprule
        \multicolumn{5}{c}{\textbf{En$\rightarrow$De } }\\
    
    $\alpha\%$ &  en$\rightarrow$nl & en$\rightarrow$et & en$\rightarrow$ru & en$\rightarrow$zh \\
    \midrule
       1\% & 12.6\textsubscript{0.2} & {7.0\textsubscript{0.7}} & {7.0\textsubscript{0.6}} & {6.7\textsubscript{0.5}}  \\
       10\% & {22.7\textsubscript{0.2}} & {12.3\textsubscript{0.3}}  &{12.7\textsubscript{0.3}}  & {13.5\textsubscript{0.3}}  \\
       50\% & {25.5\textsubscript{0.2}} & {16.0\textsubscript{0.4}}  & {17.8\textsubscript{0.2}} &{16.7\textsubscript{0.4}}  \\
       100\% & {28.4\textsubscript{0.3}}  &{16.5\textsubscript{0.1}}& {{18.2}\textsubscript{0.3}}   &  {16.5\textsubscript{0.2}} \\
       200\%  & {29.4\textsubscript{0.0}} & {15.0\textsubscript{0.0}} & {18.1\textsubscript{0.0}} & {16.4\textsubscript{0.0}} \\
    \bottomrule

    \end{tabular}}
    
       \end{subtable}
\begin{subtable}{\linewidth}
\centering
   \resizebox{0.8\linewidth}{!}{
    \begin{tabular}{r|rrrr}
    \toprule
        \multicolumn{5}{c}{\textbf{En$\rightarrow$Ru  } }\\
    
    $\alpha\%$  & en$\rightarrow$uk & en$\rightarrow$cs & en$\rightarrow$de & en$\rightarrow$zh \\
    \midrule
       1\%   & 13.8\textsubscript{0.6} &  8.2\textsubscript{0.6} & 7.0\textsubscript{0.2} & {5.8}\textsubscript{0.2}   \\
       10\%  & 18.0\textsubscript{0.5} & {11.2}\textsubscript{0.2}  & 12.5\textsubscript{0.3}& 12.3\textsubscript{0.1} \\
       50\% & 20.3\textsubscript{0.3}  & 12.6\textsubscript{0.1}  & 16.0\textsubscript{0.2} & 16.5\textsubscript{0.4} \\
       100\% & 23.7\textsubscript{0.2}& {{15.2}\textsubscript{0.1}}   &  {17.8\textsubscript{0.2}} & 16.1\textsubscript{0.1}\\
       200\%  & {26.4\textsubscript{0.0}} &{16.7}\textsubscript{0.0} &{19.9\textsubscript{0.0}} & 16.2\textsubscript{0.0} \\
    \bottomrule

    \end{tabular}}
   
       \end{subtable}

        \begin{subtable}{\linewidth}
        \centering
   \resizebox{0.8\linewidth}{!}{
    \begin{tabular}{r|rrrr}
    \toprule
        \multicolumn{5}{c}{\textbf{En$\rightarrow$Es } }\\
    
    $\alpha\%$   & en$\rightarrow$pt & en$\rightarrow$nl & en$\rightarrow$ru & en$\rightarrow$zh \\
    \midrule
       1\% &  {12.2\textsubscript{0.4}} &  {10.6\textsubscript{0.6}} & {7.2\textsubscript{0.3}} & {6.1\textsubscript{0.8}}   \\
       10\%  & {19.3\textsubscript{0.3}}  & {12.3\textsubscript{0.6}}  & {13.0\textsubscript{0.2}} & {{14.2}\textsubscript{0.3}} \\
       50\%  & {22.5\textsubscript{0.3}}  & {19.2\textsubscript{0.3}} & {17.5\textsubscript{0.3}} & {{16.5}\textsubscript{0.2}} \\
       100\% & {27.2\textsubscript{0.2}}& {{20.2}\textsubscript{0.3}}   &  {18.5\textsubscript{0.3}} & {16.6\textsubscript{0.2}}\\
       200\%  &  {28.2\textsubscript{0.0}} & {20.2\textsubscript{0.0}} & {18.6\textsubscript{0.0}} & {16.0\textsubscript{0.0}} \\
    \bottomrule

    \end{tabular}}
    
       \end{subtable}

    \caption{BLEU scores for the auxiliary language pairs in a mid-resource setting (1M) when training with main language pairs: En$\rightarrow$De, En$\rightarrow$Es, and En$\rightarrow$Ru. $\alpha\%=1, 10, 50, 100, 200$ represents the proportion of the medium-resource setting size.  }
    \label{tab:mid-resource-aux}

\end{table}

\end{document}